\documentclass{article}

\usepackage[preprint]{neurips_2026}

\usepackage[utf8]{inputenc} 
\usepackage[T1]{fontenc}    
\usepackage{hyperref}       
\usepackage{url}            
\usepackage{booktabs}       
\usepackage{amsfonts}       
\usepackage{nicefrac}       
\usepackage{microtype}      
\usepackage{xcolor}         
\usepackage{amsmath}
\usepackage{algorithm}
\usepackage{enumitem}
\usepackage{picinpar}

\usepackage{graphicx}
\usepackage{algorithmicx}
\usepackage[noend]{algpseudocode}
\usepackage{subfigure}
\usepackage{multirow}
\usepackage{color}
\usepackage{balance}
\usepackage{enumitem}
\usepackage{hhline}
\usepackage[normalem]{ulem}
\usepackage{booktabs}
\usepackage{wrapfig}
\usepackage{cancel}
\usepackage{hyperref}
\usepackage{makecell}
\usepackage{longtable}

\usepackage{capt-of}

\newcommand{\bV}{\ensuremath{\mathcal{V}}}
\newcommand{\bG}{\ensuremath{\mathcal{G}}}

\newcommand{\bN}{\ensuremath{\mathcal{N}}}
\newcommand{\bE}{\ensuremath{\mathcal{E}}}

\renewcommand{\vec}[1]{\ensuremath{\mathbf{#1}}}

\newcommand{\stitle}[1]{\vspace{0mm} \noindent {\bf #1}}

\newcommand{\method}[1]{\textsc{#1}}
\newcommand{\model}{\method{GraphReAct}{}}

\newcommand{\eat}[1]{}

\newcommand{\stkout}[1]{\ifmmode\text{\sout{\ensuremath{#1}}}\else\sout{#1}\fi}

\title{\model: Reasoning and Acting for Multi-step Graph Inference}

\author{%
\textbf{Xingtong Yu\textsuperscript{1}
\quad Zhongwei Kuai\textsuperscript{2}
\quad Chang Zhou\textsuperscript{2}
\quad Xuanting Xie\textsuperscript{3}}
\quad {Renhe Jiang\textsuperscript{4}}  \\
\textbf{Xikun Zhang\textsuperscript{5}
\quad Hong Cheng\textsuperscript{1}
\quad Xinming Zhang\textsuperscript{2}
\quad Yuan Fang\textsuperscript{6}} \\
\textsuperscript{1}The Chinese University of Hong Kong 
\quad \textsuperscript{2}The University of Science and Technology of China\\
\textsuperscript{3}University of Electronic Science and Technology of China \quad
\textsuperscript{4}The University of Tokyo\\
\textsuperscript{5}RMIT University \quad
\textsuperscript{6}Singapore Management University\\
}

\begin{document}

\maketitle

\begin{abstract}
Reasoning-acting frameworks enhance large language models (LLMs) by interleaving reasoning with actions for dynamic information acquisition. However, extending this paradigm to graph learning remains underexplored. Graph data is inherently structured, with information distributed across nodes and edges and encoded through both topology and latent representations. As a result, effective reasoning over graphs requires not only retrieving informative evidence from the graph, but also progressively refining the accumulated context during multi-step inference.
In this work, we propose \model, a graph reasoning-acting framework that enables step-by-step inference over graph-structured data. Specifically, we design a graph-based action space with two complementary retrieval actions: topological retrieval, which captures local structural dependencies, and semantic retrieval, which accesses non-local but relevant evidence in the representation space. These actions dynamically expand the reasoning context. To further support multi-step reasoning, we introduce another type of action, context refinement, which distills and reorganizes accumulated information into a compact representation. By interleaving reasoning with both retrieval and refinement actions, our framework enables a progressive transition from context expansion to compression.
Extensive experiments on six benchmark datasets demonstrate that \model\ consistently outperforms state-of-the-art methods, validating the effectiveness of reasoning-acting for graph learning. 
\end{abstract}

\section{Introduction}
\begin{figure}[t]
\centering
\includegraphics[width=1\linewidth]{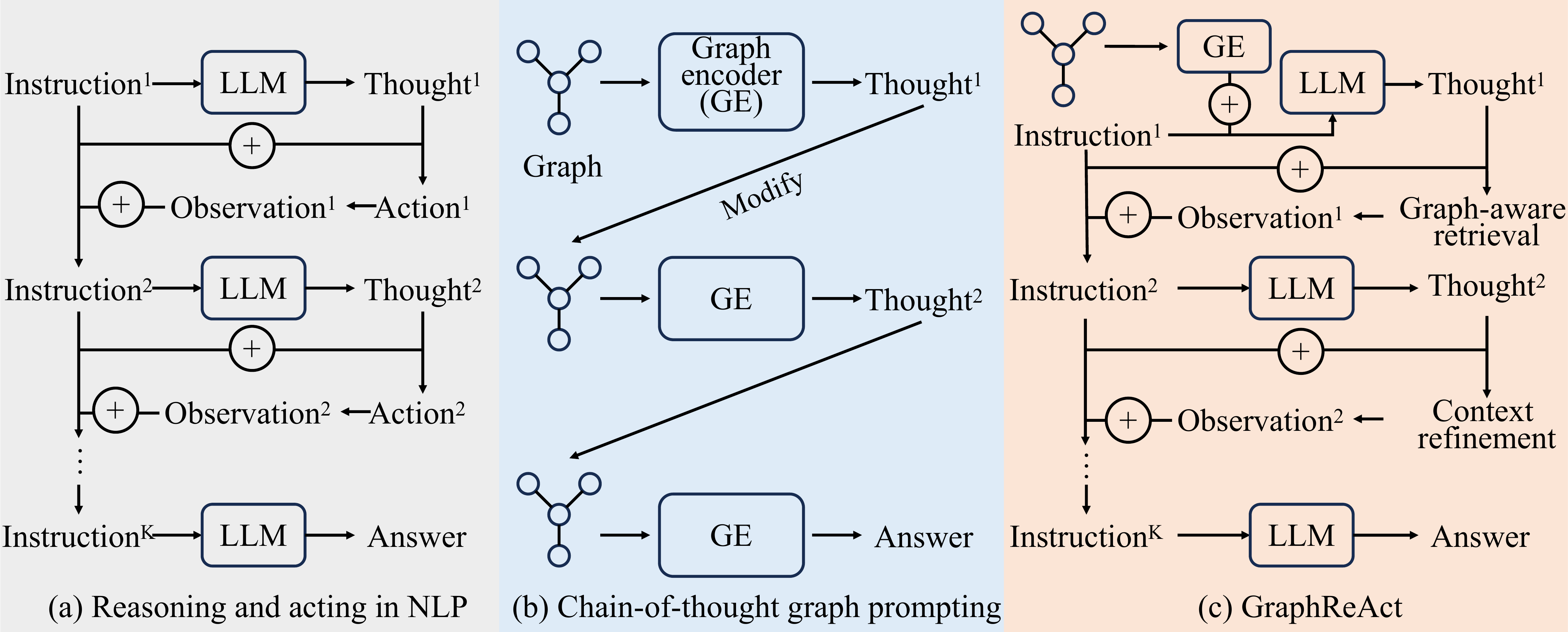}
\vspace{-6mm}
\caption{Comparison of reasoning paradigms. (a) Reasoning and acting in NLP. (b) CoT-based graph prompting. (c) Our proposed graph reasoning-acting (\model) framework.}
\label{fig.intro}
\end{figure}

Recent advances in large language models have demonstrated strong reasoning capabilities through Chain-of-Thought (CoT) prompting \citep{wei2022chain,feng2023towards,zhangautomatic}, which decomposes complex problems into intermediate reasoning steps. Building upon this paradigm, the reasoning-acting framework \citep{yao2022react,shensatori,fu2025preact} further enhances reasoning by interleaving it with actions, enabling models to interleave reasoning with structured actions to acquire additional information during inference, as shown in Fig.~\ref{fig.intro}(a). This reasoning-acting synergy has proven particularly effective in tasks such as question answering and decision making, where actions (e.g., search) help reduce hallucination and improve factual accuracy. These successes suggest that integrating reasoning with information acquisition is a powerful paradigm for solving complex problems.

Despite its effectiveness in natural language processing, extending reasoning-acting framework to graph learning remains largely underexplored. Graph-structured data \citep{xia2021graph,cook2006mining} is inherently different from textual environments: information is distributed across nodes and edges, and meaningful signals are often encoded implicitly through topology and latent representations \citep{kipf2016semi,velivckovic2017graph}. Existing graph learning methods, including end-to-end graph encoder \citep{xu2018powerful,hamilton2017inductive}, finetuning \citep{you2020graph,velivckovic2018deep} and graph prompting \citep{liu2023graphprompt,yu2023generalized} approaches, typically rely on a fixed receptive field and perform single-pass inference, which may limit their ability to resolve ambiguity under sparse or incomplete information. While a recent work has incorporated CoT-style reasoning into graph-related tasks \citep{yu2025gcot}, it lacks an action mechanism to dynamically expand the available context, as shown in Fig.~\ref{fig.intro}(b). A key obstacle is the mismatch between the unstructured action space in natural language domain (e.g., web search \citep{yao2022react}) and the structured nature of graph data. This raises a fundamental question: \textit{how can we enable reasoning-acting synergy in graph learning through appropriate action design?} This problem is non-trivial due to two key challenges.

First, in natural language domain, actions typically correspond to interactions with external knowledge sources such as Wikipedia \citep{yao2022react,fu2025preact}. In contrast, for graph learning tasks, while external information remains accessible, it is often misaligned with the task objective and may introduce irrelevant or noisy signals. Instead, the most informative evidence is inherently structured within the graph itself, encoded through both topology and latent representations. Therefore, directly transferring text-based action designs to graph domains is suboptimal. This raises a key challenge: \textit{how to design graph-aware actions that can effectively retrieve informative evidence from both topological structure and semantic similarity?}  
In this work, we propose a graph-based retrieval action space that enables structured information acquisition from the graph, as shown in Fig.~\ref{fig.intro}(c). Specifically, we design two complementary actions: (1) \emph{topological retrieval}, which gathers textual information from structurally neighboring nodes to capture local dependencies, and (2) \emph{semantic retrieval}, which identifies nodes with similar embeddings to provide non-local, semantically relevant evidence. By jointly leveraging these two actions, our framework effectively balances locality and globality, allowing the model to dynamically expand the context beyond fixed receptive fields.

Second, \textit{how to support multi-step reasoning with actions in graph domains?} In standard reasoning-acting frameworks, actions are typically used for information acquisition, where the model queries external environments to gather additional evidence \citep{yao2022react}.
However, graph reasoning exhibits a different requirement. While early-stage reasoning benefits from expanding the context through information retrieval, later stages require consolidating and distilling the accumulated information to avoid redundancy and noise. This introduces an inherent tension between context expansion and context compression, which is not explicitly addressed in existing reasoning-acting frameworks. To address this challenge, we design another type of action, termed \emph{context refinement}, tailored for graph-aware LLM reasoning. Specifically, after the initial graph-aware retrieval stage, subsequent actions focus on context refinement, where the model distills, summarizes, and reorganizes the accumulated information into a more compact and informative representation. This design enables a natural transition from expansion to compression, allowing the model to perform multi-step reasoning over a progressively constructed context, rather than relying on repeated retrieval from the graph.
Notably, unlike fully agentic reasoning-acting frameworks, our action sequence follows a structured design tailored to graph data, where retrieval and refinement are applied in a predefined order to ensure stable and efficient reasoning.

In summary, our contributions are fourfold.
(1) We propose \model, a novel reasoning-acting framework for graph learning that enables step-by-step inference via structured interaction with graph-structured data. To the best of our knowledge, this is the first work that systematically extends the reasoning-acting paradigm to graph domains.
(2) We propose a graph-based retrieval action space with topology-based and semantic retrieval, enabling structured and complementary information acquisition from both local neighborhoods and global semantic space.
(3) We design context refinement action that balances context expansion and compression, enabling progressive refinement of graph-aware information during inference.
(4) We conduct extensive experiments on six datasets, demonstrating that \model\ consistently outperforms state-of-the-art methods and validating the effectiveness of reasoning-acting in graph learning.

\section{Related Work}

\stitle{LLMs for graph learning.}
Recent studies integrate LLMs into graph learning in three main ways. Some use LLMs as semantic feature extractors to enrich node representations and mitigate domain gaps \citep{wang2024gft,he2025unigraph2,liu2023one,yuan2025graver}. Others employ LLMs as predictors to leverage their reasoning capabilities \citep{tang2024graphgpt,tang2024higpt,chen2024llaga}. There are also efforts to align graph representations with LLM embedding spaces to enable graph--language interaction \citep{tang2024graphgpt,chen2024llaga}. Despite these advances, most methods rely on static graph--language interfaces with single-pass inference, lacking explicit multi-step reasoning and structured evidence acquisition. In contrast, our method enables iterative evidence integration through graph-aware operations, supporting multi-step reasoning over graph data.

\stitle{Reasoning and acting in LLMs.}
Chain-of-Thought (CoT) prompting improves reasoning by decomposing problems into intermediate steps \citep{wei2022chain,zhou2022least,wang2022self,yao2023tree}. Building on this, reasoning-acting frameworks interleave reasoning with external actions to acquire additional information \citep{yao2022react,fu2025preact}. However, directly applying such approaches to graph learning is non-trivial, as graph information is inherently structured and distributed across topology and latent semantic space. Existing methods neither exploit these structured inductive biases nor address the need to organize evidence acquisition and refinement in multi-step reasoning. Our work instead introduces a structured mechanism tailored to graph data, combining graph-based retrieval with progressive context refinement.
\section{Preliminaries}\label{sec.prelim}

\stitle{Graph encoder.} A graph is defined as \(G=(\bV ,\bE ,\vec{X})\), where $\bV$ and $\bE$ denote the sets of nodes and edges, respectively, and $\vec{X}$ is the node feature matrix, whose $i$-th row $x_i$ corresponds to the feature vector of node \(v_i \in \bV\). We denote a collection of graphs as $\bG$.
A mainstream technique for graph representation learning is GNNs, which recursively update node representations through message passing. 
Let \( \vec{H}^l \in \mathbb{R}^{|\bV| \times d} \) denote the embedding matrix at the $l$-th layer, where the $i$-th row \( \vec{h}_i^l \) represents the embedding of node $v_i$. The layer-wise update is defined as
\begin{align}
   \vec{h}^l_v = \mathtt{MP}(\vec{h}^{l-1}_v, \{\vec{h}^{l-1}_u : u\in\bN_v\}; \theta^l),
\end{align}
where $\bN_v$ denotes the neighbors of $v$, $\mathtt{MP}$ denotes the message-passing function, and $\theta^l$ represents the learnable parameters of the $l$-th layer. The initial embedding is given by $\vec{h}^0_v = x_v$, and after $L$ layers, the final node representations are denoted as 
\begin{align}
    \vec{H} = \mathtt{GE}(\vec{X}, G; \Theta),
\end{align}
where \(\Theta = \{\theta^{1},\dots,\theta^{L}\}\).

\stitle{Graph--LLM interface.}
Recent approaches introduce a graph-to-language interface that integrates graph encoders with LLMs, enabling LLMs to perform prediction over graph-structured data. 
To bridge the gap between graph representations and the LLM token space, existing methods typically first conduct pre-training based on multimodal contrastive learning \citep{wang2024llms,chen2024llaga}. Given a graph $G$ and its associated text, the graph encoder $\mathtt{GE}$ is trained to produce representations that are aligned with the embedding space of the LLM. Specifically, the pre-training objective encourages graph representations to align with corresponding textual representations, thereby embedding graph features into the LLM semantic space. This alignment enables graph embeddings to be interpreted as pseudo-tokens that are compatible with LLM inputs.
After pre-training, the graph encoder is frozen. Given an input graph, it first produces representations $\vec{H}_\bV = \mathtt{GE}(\vec{X}, G; \Theta)$, and then maps them into the token embedding space of the LLM via a projection function:
\begin{align}
    \vec{H}^{\mathrm{tok}} = \mathtt{Proj}(\vec{H};\phi),
\end{align}
where $\phi$ is the learnable parameter. The graph embedding tokens $\vec{H}^{\mathrm{tok}}$ share the same embedding dimension as LLM tokens and are incorporated into an instruction template for the LLM.
\section{Proposed Approach}
In this section, we introduce \model. We first provide a high-level overview of the framework and then describe its core components in detail.

\subsection{Overall Framework}

\begin{figure}[t]
\centering
\includegraphics[width=1\linewidth]{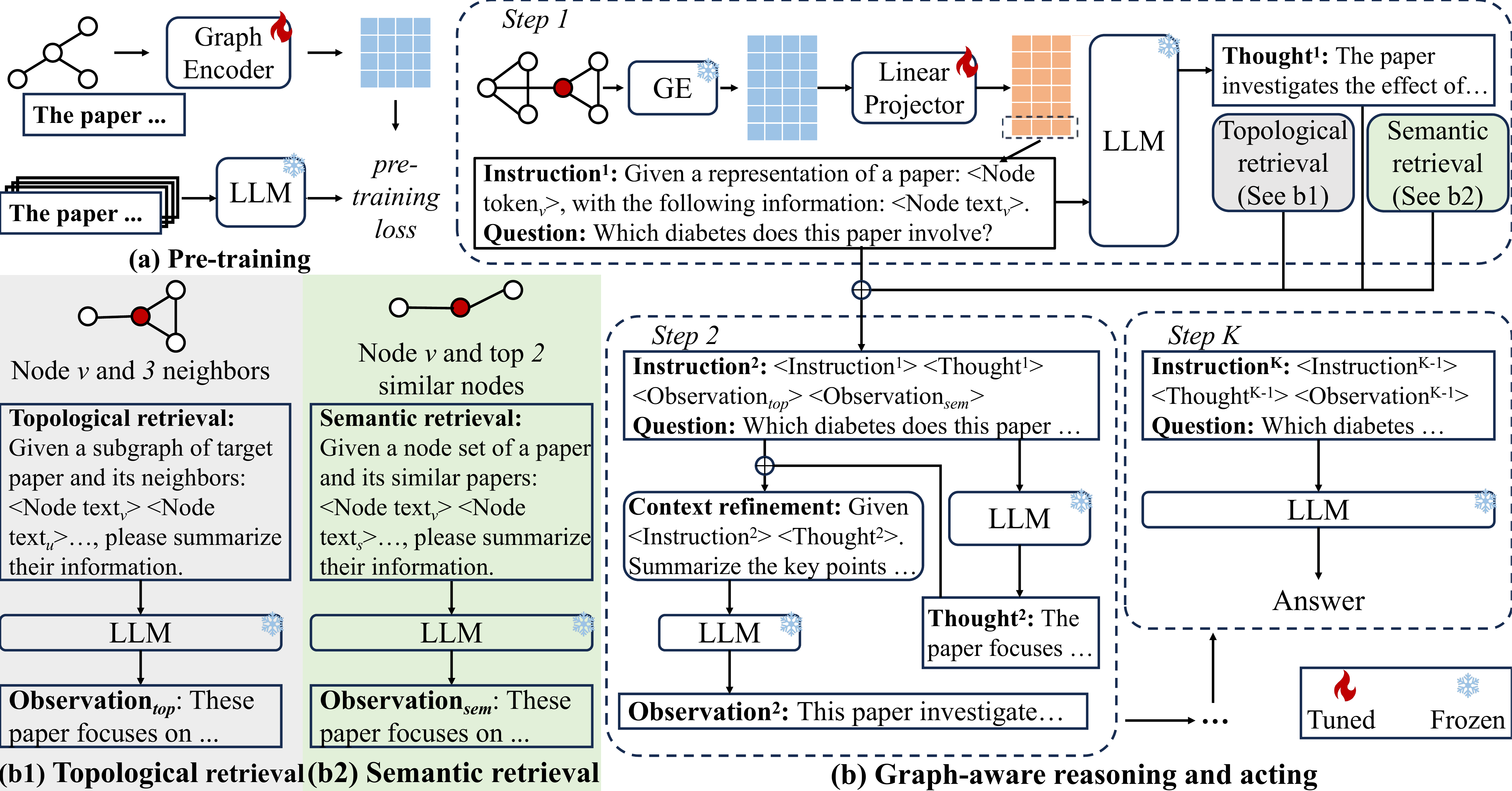}
\vspace{-6mm}
\caption{Overall framework of \model.}
\label{fig.framework}
\end{figure}

We illustrate the overall framework of \model\ in Fig.~\ref{fig.framework}, which consists of two phases: (1) pre-training, and (2) graph-based reasoning and acting. 
First, we pre-train a graph encoder to produce representations aligned with the embedding space of a large language model, as shown in Fig.~\ref{fig.framework}(a). Details of the pre-training phase are provided in Sect.~\ref{sec.prelim}.
Second, we propose a graph-based reasoning and acting framework to guide the LLM to perform multi-step inference via iterative context updating, as illustrated in Fig.~\ref{fig.framework}(b). Specifically, at each step, the LLM generates an intermediate thought based on the current context, which is used to guide a predefined action for updating the context.
We consider two types of actions. The first type is graph-based retrieval, which expands the context by incorporating information from topologically related nodes and semantically similar nodes, as shown in Fig.~\ref{fig.framework}(b1,b2). The second type is context refinement, which compresses and reorganizes the accumulated context to reduce redundancy and enhance relevance. Through this progressive context updating process, the model constructs a high-quality context for final prediction.

\subsection{Graph-based Retrieval}
Graph-based retrieval is designed to acquire informative evidence from graph-structured data, where useful signals reside in both graph topology and latent representation space. To this end, we design a unified set of action primitives that captures two complementary sources of information: structural locality and semantic similarity.

\stitle{Topological retrieval.}
We first introduce a topology-based action that extracts structurally grounded evidence from the graph. In graphs, informative signals are often distributed along connectivity patterns, making local neighborhoods a primary source of context for reasoning.
To capture such structural dependencies, we retrieve nodes that are topologically related to the target node $v$ via a breadth-first traversal \citep{bundy1984breadth}, collecting the first $N$ visited neighbors. Unlike random neighbor sampling, this traversal expands the receptive field in a structured manner, ensuring a consistent retrieval budget across nodes while adaptively incorporating higher-order neighborhoods when necessary. 

We then transform the retrieved structural evidence into a compact representation suitable for reasoning. Specifically, we construct a topology-based instruction by integrating the target node text with the textual information of the retrieved neighbors, and use the frozen LLM to generate a topology-based summary. This summary, denoted as $s^{\mathrm{top}}$, serves as a structured abstraction of the local neighborhood, encoding aggregated structural signals into a form that can be directly consumed by subsequent reasoning steps.

\stitle{Semantic retrieval.}
While topological retrieval captures local structural dependencies, it may fail to access semantically relevant but structurally distant evidence. To address this limitation, we introduce a complementary semantic retrieval action that operates in the representation space.
Specifically, we identify nodes that are semantically similar to the target node $v$ based on cosine similarity \citep{xia2015learning} between node embeddings, and select the top-$M$ most similar nodes. This process enables the model to access globally relevant evidence that is not constrained by graph connectivity, thereby extending the reasoning context beyond local neighborhoods.

Similar to topological retrieval, we convert the retrieved semantic evidence into a reasoning-compatible representation. We construct a semantic-based instruction by integrating the target node text with the textual information of the retrieved semantically similar nodes, and feed it into the frozen LLM to produce a semantic-based summary. This summary, denoted as $s^{\mathrm{sem}}$, provides a global semantic abstraction of the target node, complementing the locality of topology-based context and enabling reasoning over both structural and semantic dimensions.
TThis dual-retrieval design provides a structured mechanism for accessing both local and global evidence, forming the foundation for graph-aware reasoning in subsequent steps. 

\stitle{Comparison with heterophilic graph learning.}
Existing methods for modeling non-local dependencies, particularly in heterophilic graphs \citep{ma2021homophily,luan2022revisiting}, typically extend message passing with multi-hop aggregation or similarity-based propagation. However, these approaches rely on fixed aggregation schemes learned during training, and rely on fixed aggregation schemes and lack instance-specific evidence selection at inference time. Moreover, they operate in a single-modality setting and are not designed for zero-shot generalization across datasets. In contrast, our approach formulates graph information access as an explicit action in the reasoning process. Instead of static propagation, our approach performs explicit evidence acquisition via topological and semantic retrieval, and converts them into natural language summaries for subsequent reasoning. This enables flexible and instance-specific evidence integration, which is particularly suitable for zero-shot graph learning.

\stitle{Context construction.}
Graph-based retrieval are applied after the initial reasoning step to construct an evidence-augmented context for subsequent reasoning.
We first perform an initial inference using the graph--LLM interface:
\begin{align}
\text{Thought}^1 = \mathtt{LLM}(\mathcal{I}, \mathcal{Q}),
\end{align}
where $\mathcal{I}$ denotes the initial instruction formed by combining the target node text $x^{\mathrm{text}}$ and its node token embedding $\mathbf{h}_V^{\mathrm{tok}}$, and $\mathcal{Q}$ denotes a task-level instruction specifying the prediction objective. $\mathcal{Q}$ is shared across all nodes and remains fixed during inference.
We then retrieve structured evidence from the graph via graph-based retrieval:
\begin{align}
\text{Observation}^1 = \mathtt{Act}_{\text{retrieve}}(G, v) = \{s^{\mathrm{top}},\; s^{\mathrm{sem}}\},
\end{align}
where $s^{\mathrm{top}}$ and $s^{\mathrm{sem}}$ denote the topological and semantic summaries, respectively. $\mathtt{Act}_{\text{retrieve}}(\cdot)$ denotes a composite operation that includes topological and semantic retrieval and  LLM-based abstraction into textual summaries.
Finally, we construct the initial context by integrating the initial thought with the retrieved evidence:
\begin{align}\label{eq.c1}
\mathcal{C}^{1} = \mathtt{Init}(\text{Thought}^1,\; \text{Observation}^1),
\end{align}
where $\mathtt{Init}(\cdot)$ denotes a prompt construction operation that organizes the thought and retrieved summaries into a structured instruction that serves as the input context for the LLM. 
$\mathcal{C}^{1}$ corresponds to a structured textual instruction that encodes accumulated evidence and reasoning states fed into the LLM. This initial context serves as the starting point for subsequent multi-step reasoning, where the context is progressively refined through iterative updates.
Notably, these retrieval operations are applied in a predefined manner rather than being dynamically selected by the model, reflecting the structured nature of graph data.
We provide detailed instruction templates in Appendix~\ref{app.sec.prompt}.

\subsection{Context Refinement}
Context refinement is designed to support multi-step reasoning by progressively updating the reasoning context under the guidance of intermediate thoughts. Unlike graph-based retrieval, which acquires external evidence from the graph, refinement is formulated as another type of action within the same action space, operating on the accumulated context and reasoning signals to enable structured consolidation and reorganization of information across steps.

At each reasoning step, the model first generates a new thought based on the current context. This thought is then used to guide a predefined refinement operation that updates the context. To unify different forms of context updates, we define a generalized refinement operator that produces an observation:
\begin{align}\label{eq.c2}
\text{Observation}^{k+1} = \mathtt{Act}_{\text{refine}}(\mathcal{C}^{k},\; \text{Thought}^{k+1}),
\end{align}
where $\mathtt{Act}_{\text{refine}}(\cdot)$ denotes a refinement action that transforms the accumulated context into a more compact and informative representation.
The updated context is then constructed by integrating the previous context, the newly generated thought, and the resulting observation:
\begin{align}\label{eq.c3}
\mathcal{C}^{k+1} = \mathtt{Update}(\mathcal{C}^{k},\; \text{Thought}^{k+1},\; \text{Observation}^{k+1}),
\end{align}
where $\mathtt{Update}(\cdot)$ incorporates the observation into the existing context while preserving relevant historical information.
Specifically, the refinement action is implemented via instruction-guided generation, where the LLM performs a reasoning-conditioned transformation over $\mathcal{C}^{k}$ and $\text{Thought}^{k+1}$, conducting information distillation and structural reorganization. The resulting observation serves as a condensed abstraction of the accumulated context, analogous to the evidence summaries obtained during graph-based retrieval.

To summarize the overall reasoning process, at each step $k$, the LLM operates on a structured input consisting of the node-specific instruction, current context, and task query:
\begin{align}
\text{Thought}^{k+1} = \mathtt{LLM}(\mathcal{I},\; \mathcal{C}^{k},\; \mathcal{Q}).
\end{align}
The generated thought is then used to guide the corresponding refinement operation as defined in Eq.~(\ref{eq.c2}) and Eq.~(\ref{eq.c3}). 
At the final step, the LLM directly produces the prediction:
\begin{align}
\hat{y} = \mathtt{LLM}(\mathcal{I},\; \mathcal{C}^{K},\; \mathcal{Q}).
\end{align}

\subsection{Adaptation and Inference}
We consider a node classification task with a labeled training set $\mathcal{D} = \{(v_1, y_1), (v_2, y_2), \dots\},$ where \( v_i \in \bV \) denotes a target node and \( y_i \in Y \) is its class label.
Given a target node \( v \), our framework performs multi-step reasoning over the constructed context and produces a predictive distribution over labels via the LLM. Let \( p(y \mid v) \) denote the probability of predicting label \( y \) from the LLM output conditioned on the final instruction. The training objective is defined as the negative log-likelihood:
\begin{align}
\mathcal{L}_{\text{down}}(\mathcal{D};\phi)
=
- \sum_{(v_i, y_i) \in \mathcal{D}}
\log p(y_i \mid v_i;\phi),
\end{align}
where the probability is derived from the LLM output via label verbalization or answer matching.
During training, both the graph encoder and the LLM are kept frozen. The only trainable component is the projection function \( \mathtt{Proj}(\cdot;\phi) \), which maps graph representations into the LLM token embedding space. By optimizing \( \phi \), the model learns to better align graph-derived representations with the LLM input space for downstream prediction.

For cross-dataset inference, the learned projection function can be directly applied to unseen graphs without further training. Specifically, given a new graph and a target instance, we construct the corresponding instruction using the projected graph representations, and perform the same multi-step reasoning process to obtain predictions. This enables \model\ to be applied to unseen graphs in a zero-shot manner, leveraging the learned graph--LLM alignment without requiring additional task-specific training.

\section{Experiments}
In this section, we conduct experiments to evaluate \model\ and analyze the empirical results. 

\subsection{Experimental Setup}

\stitle{Datasets.}
We evaluate \model\ on eight text-attributed graph datasets from two domains: citation networks and e-commerce networks. 
For citation networks, we use Arxiv~\cite{hu2020open}, PubMed~\cite{he2024harnessing}, and an expanded version of Cora~\cite{wen2023augmenting}, where nodes represent papers and edges denote citation relationships. 
For e-commerce networks, we use Computer, Photo, Children, History, and Sports from the TAG benchmark~\cite{yan2023comprehensive}, where nodes represent products and edges indicate frequent co-viewing or co-purchasing relationships. 
Detailed statistics of all datasets are provided in Table~\ref{tab:dataset}, and further descriptions are provided in Appendix~\ref{app:dataset_description}.

\begin{table}[b]
  \centering
  \caption{Accuracy of zero-shot node classification.}
  \label{tab:acc}
  \setlength{\tabcolsep}{6pt}
  \resizebox{\linewidth}{!}{
  \begin{tabular}{lcc|cccc}
    \toprule
    Model & \multicolumn{2}{c|}{Citation} & \multicolumn{4}{c}{E-commerce} \\
    \cmidrule(r){2-3} \cmidrule(l){4-7}
    & Cora & Pubmed & Children & History & Photo & Sports \\
    \midrule
    \method{MLP}          & 0.021$\pm$0.006 & 0.323$\pm$0.027 & 0.029$\pm$0.037 & 0.080$\pm$0.041 & 0.110$\pm$0.070 & 0.042$\pm$0.021 \\
    \midrule
    \method{GCN}          & 0.017$\pm$0.004 & 0.288$\pm$0.092 & 0.030$\pm$0.018 & 0.063$\pm$0.042 & 0.103$\pm$0.047 & 0.042$\pm$0.025 \\
    \method{GraphSAGE}    & 0.014$\pm$0.007 & 0.316$\pm$0.058 & 0.008$\pm$0.007 & 0.195$\pm$0.206 & 0.056$\pm$0.055 & 0.051$\pm$0.015 \\
    \method{GAT}          & 0.016$\pm$0.004 & 0.343$\pm$0.064 & 0.086$\pm$0.084 & 0.172$\pm$0.098 & 0.050$\pm$0.027 & 0.142$\pm$0.138 \\
    \method{NodeFormer}   & 0.016$\pm$0.007 & 0.308$\pm$0.093 & 0.048$\pm$0.028 & 0.168$\pm$0.127 & 0.073$\pm$0.015 & 0.165$\pm$0.057 \\
    \method{DIFFormer}    & 0.029$\pm$0.014 & 0.361$\pm$0.071 & 0.129$\pm$0.030 & 0.275$\pm$0.171 & 0.321$\pm$0.055 & 0.306$\pm$0.131 \\
    \midrule
    \method{DGI}          & 0.026$\pm$0.009 & 0.329$\pm$0.103 & 0.082$\pm$0.035 & 0.218$\pm$0.168 & 0.224$\pm$0.127 & 0.049$\pm$0.017 \\
    \midrule
    \method{GKD}          & 0.042$\pm$0.008 & 0.399$\pm$0.033 & 0.202$\pm$0.064 & 0.339$\pm$0.138 & 0.166$\pm$0.086 & 0.208$\pm$0.077 \\
    \method{GLNN}         & 0.031$\pm$0.006 & 0.390$\pm$0.011 & 0.187$\pm$0.012 & 0.283$\pm$0.021 & 0.403$\pm$0.019 & 0.317$\pm$0.048 \\
    \midrule
    \method{Vicuna-7B-v1.5}  & 0.156$\pm$0.001 & 0.719$\pm$0.010 & 0.270$\pm$0.001 & 0.363$\pm$0.001 & 0.378$\pm$0.004 & 0.370$\pm$0.001 \\
    \method{Vicuna-7B-SPT}   & 0.168$\pm$0.018 & 0.768$\pm$0.036 & 0.227$\pm$0.015 & 0.281$\pm$0.088 & 0.350$\pm$0.061 & 0.230$\pm$0.018 \\
    \midrule
    \method{OFA}          & 0.130$\pm$0.019 & 0.314$\pm$0.059 & 0.064$\pm$0.086 & 0.052$\pm$0.049 & 0.340$\pm$0.026 & 0.101$\pm$0.071 \\
    \method{GraphGPT-std} & 0.126 & 0.701 & --- & --- & --- & --- \\
    \method{GraphGPT-cot} & 0.181 & 0.521 & --- & --- & --- & --- \\
    \method{LLaGA}        & 0.168$\pm$0.032 & 0.793$\pm$0.036 & 0.199$\pm$0.007 & 0.146$\pm$0.067 & 0.276$\pm$0.069 & 0.352$\pm$0.033 \\
    \method{TEA-GLM}      & \underline{0.202}$\pm$0.014 & \textbf{0.848}$\pm$0.010 & \underline{0.271}$\pm$0.010 & \underline{0.528}$\pm$0.058 & \underline{0.497}$\pm$0.027 & \underline{0.404}$\pm$0.010 \\
    \model      & \textbf{0.273}$\pm$0.002 & \underline{0.819}$\pm$0.001 & \textbf{0.294}$\pm$0.001 & \textbf{0.645}$\pm$0.001 & \textbf{0.523}$\pm$0.003 & \textbf{0.483}$\pm$0.001 \\
    \bottomrule
  \end{tabular}
  }
  
\vspace{1mm}
  \parbox{\linewidth}{\footnotesize \ \ The best method is bolded and the runner-up is underlined.}
\end{table}
\stitle{Baselines.}
We compare \model\ with representative methods across five categories.
(1) \emph{Non-graph model}: 
MLP \citep{taud2017multilayer} serves as a structure-agnostic baseline that relies solely on node features without exploiting graph topology.
(2) \emph{Supervised graph methods}: 
GCN~\citep{kipf2016semi}, \method{GraphSAGE}~\citep{hamilton2017inductive}, and GAT~\citep{velivckovic2017graph} perform supervised learning via message passing over graph structures, while \method{NodeFormer}~\citep{wu2022nodeformer} and \method{DIFFormer}~\citep{wu2023difformer} extend transformer architectures to graphs to capture long-range dependencies.
(3) \emph{Self-supervised graph methods}: 
DGI~\citep{velivckovic2018deep} learns node representations via contrastive objectives on unlabeled graphs, followed by a classifier for downstream prediction.
(4) \emph{Graph knowledge distillation}: 
GKD~\citep{yang2022geometric} transfers structural knowledge from a teacher GNN trained on a full graph to a student model operating under restricted structural access, while GLNN~\citep{zhang2022graphless} distills graph-aware representations into an MLP-like architecture to reduce reliance on graph connectivity during inference.
(5) \emph{Large language models}: 
\method{Vicuna-7B-v1.5}~\citep{chiang2023vicuna} and \method{Vicuna-7B-SPT} evaluate the capability of LLMs for graph tasks via textual prompting without explicit graph modeling.
(6) \emph{Graph--LLM methods}: 
OFA~\citep{liu2023one}, \method{GraphGPT}~\citep{tang2024graphgpt}, \method{LLaGA}~\citep{chen2024llaga}, and TEA-GLM~\citep{wang2024llms} integrate graph representations with LLMs to enable zero-shot or few-shot inference on graph tasks.
Detailed descriptions of all baselines are provided in Appendix~\ref{app:baseline_description}, with implementation details in Appendix~\ref{app:implementation_details}.

\stitle{Evaluation setting.}
We follow the cross-dataset zero-shot evaluation protocol of TEA-GLM~\citep{wang2024llms}. 
Specifically, Arxiv and Computer are used as source datasets to conduct pre-training and downstream adaptation, and all methods are evaluated on unseen target datasets without further adaptation.
For the citation domain, PubMed and Cora are used as target datasets; for the e-commerce domain, we evaluate on Children, History, Photo, and Sports. 
We adapt the same data splits as TEA-GLM, i.e., 90,941 nodes for training on Arxiv and 62,748 nodes for training on Computer, with 1,000 nodes used for evaluation on the other datasets.
For all baseline methods, we directly report the results reported in TEA-GLM.
We report accuracy and Macro-F1 for node classification. 
Each experiment is conducted with five random seeds, and we report the mean and standard deviation.



\subsection{Performance Evaluation}\label{sec.performance}
We make the following observations. 
First, \model\ consistently outperforms or remains highly competitive with all baselines across datasets, demonstrating the effectiveness of incorporating graph-aware reasoning and acting into zero-shot graph learning. Compared with conventional graph learning methods, the performance gains indicate that dynamically acquiring and refining graph-aware evidence provides informative and transferable context for cross-dataset prediction. 
Second, while LLM-based and graph--LLM methods, such as LLaGA and TEA-GLM,  incorporate textual or graph representations into LLM inference, they typically rely on a fixed and static context, limiting their ability to adaptively exploit available evidence.  In contrast, \model\ introduces a dynamic reasoning process that iteratively expands and refines the context through graph-aware retrieval and context refinement actions.  This allows the model to progressively integrate topological and semantic evidence while filtering out irrelevant information, leading to more informative and task-relevant representations for prediction.
On PubMed, \model\ achieves performance comparable to TEA-GLM. One possible reason is that PubMed contains only three coarse-grained classes, where semantically or topologically similar nodes (e.g., papers on related types of diabetes) may introduce ambiguous signals during context expansion, limiting the advantage of reasoning-based refinement.

\stitle{Ablation study.}
We analyze the contribution of each component in Table~\ref{tab:ablation}. 
First, comparing Variant 1 and Variants 2--4 shows that both topological retrieval and semantic retrieval consistently improve performance over the no-retrieval baseline, demonstrating the effectiveness of graph-aware evidence acquisition. Among them,  topological retrieval brings slightly larger gains than SR, indicating the importance of structural signals.
Second, combining topological and semantic retrieval (Variant 4) further improves performance, suggesting that the two types of retrieval provide complementary information from structural and semantic perspectives.
Third, context refinement  alone (Variant 5) yields limited improvement, as it operates without additional external evidence. However, when combined with both topological and semantic retrieval, the full model achieves the best performance across all datasets, highlighting that refinement plays a crucial role in consolidating retrieved information and enhancing reasoning quality.

\begin{table}[t]
    \centering
    \caption{Ablation study of key componentss, reporting accuracy. }
    \label{tab:ablation}
    \vspace{-1mm}
    \resizebox{0.9\linewidth}{!}{
    \begin{tabular}{@{}l|ccc|cccc@{}}
    \toprule
    Methods
    & TR & SR & CF & Cora & Children & History & Sports\\
    \midrule\midrule
    \method{Variant 1}
    & $\times$ & $\times$ & $\times$ & 0.248$\pm$0.001 & 0.270$\pm$0.003 & 0.568$\pm$0.002 & 0.404$\pm$0.008\\
    \method{Variant 2}
    & $\checkmark$ & $\times$ & $\times$ & 0.274$\pm$0.003 & 0.290$\pm$0.002 & 0.603$\pm$0.006 & 0.445$\pm$0.001\\
    \method{Variant 3}
    & $\times$ & $\checkmark$ & $\times$ & 0.256$\pm$0.002 & 0.283$\pm$0.003 & 0.599$\pm$0.001 & 0.418$\pm$0.002\\
    \method{Variant 4}
    & $\checkmark$ & $\checkmark$ & $\times$ & \textbf{0.285}$\pm$0.001 & 0.293$\pm$0.004 & 0.631$\pm$0.001 & 0.421$\pm$0.001\\
    \method{Variant 5}
    & $\times$ & $\times$ & $\checkmark$ & 0.255$\pm$0.006 & 0.272$\pm$0.003 & 0.578$\pm$0.005 & 0.410$\pm$0.003\\
    \method{\model}
    & $\checkmark$ & $\checkmark$ & $\checkmark$ & 0.273$\pm$0.002 & \textbf{0.294}$\pm$0.001 & \textbf{0.645}$\pm$0.001 & \textbf{0.483}$\pm$0.001\\
    \bottomrule
    \end{tabular}}
  
\vspace{1mm}
  \parbox{0.9\linewidth}{\footnotesize \ \ TR: topological retrieval; SR: semantic retrieval; CF: context refinement.}
\end{table}

\subsection{Comparison with Textual Search Action}
We further analyze whether the standard text-based \textbf{Search} action is useful for graph reasoning. 
Unlike our graph-aware actions, \textbf{Search} retrieves external evidence from a Wikipedia-based knowledge source.
We compare two variants in Table~\ref{tab:search_action}: \method{Search}, which relies solely on external retrieval, and \model\method{+Search}, which augments \model\ with this action.
Specifically, given the current context, we first prompt the LLM to generate a query entity, then retrieve up to the top-$6$ relevant Wikipedia pages based on textual similarity, and finally concatenate the retrieved passages into the instruction as additional context for the next reasoning step.

As shown in Table~\ref{tab:search_action}, \method{Search} alone performs substantially worse than graph-based methods, indicating that external textual evidence is insufficient for node classification in graph domains. Moreover, incorporating textual search into \model\  does not lead to consistent improvements, and even slightly degrades performance compared to the full model. This suggests that external information may introduce noise or misaligned signals that interfere with graph-aware reasoning. 
We further analyze the effect of the number of retrieved entities in the search process in Appendix~\ref{app.sec.search}.
These results highlight that, unlike in NLP tasks, effective evidence for graph reasoning is primarily encoded within the graph itself, and directly applying text-based search actions is neither necessary nor beneficial.

\begin{wraptable}{r}{0.6\columnwidth}
\vspace{-2mm}
\centering
\caption{Analysis of the textual search action.}
\label{tab:search_action}
\resizebox{\linewidth}{!}{
\begin{tabular}{@{}l|cccc@{}}
\toprule
Methods & Cora & Children & History & Sports\\
\midrule
\method{Search} & 0.253 & 0.225 & 0.535 & 0.427\\
\model\method{+Search} & 0.269 & 0.282 & 0.635 & 0.479\\
\model & \textbf{0.273} & \textbf{0.294} & \textbf{0.645} & \textbf{0.483}\\
\bottomrule
\end{tabular}}
\vspace{-4mm}
\end{wraptable}

\subsection{Parameter Sensitivity Analysis}

We study the sensitivity of three key hyperparameters: the number of inference steps,the size of topological retrieved and semantic retrieval.

\begin{figure}[t]
    \centering
    \begin{minipage}[b]{0.31\linewidth}
        \centering
        \includegraphics[width=0.93\linewidth]{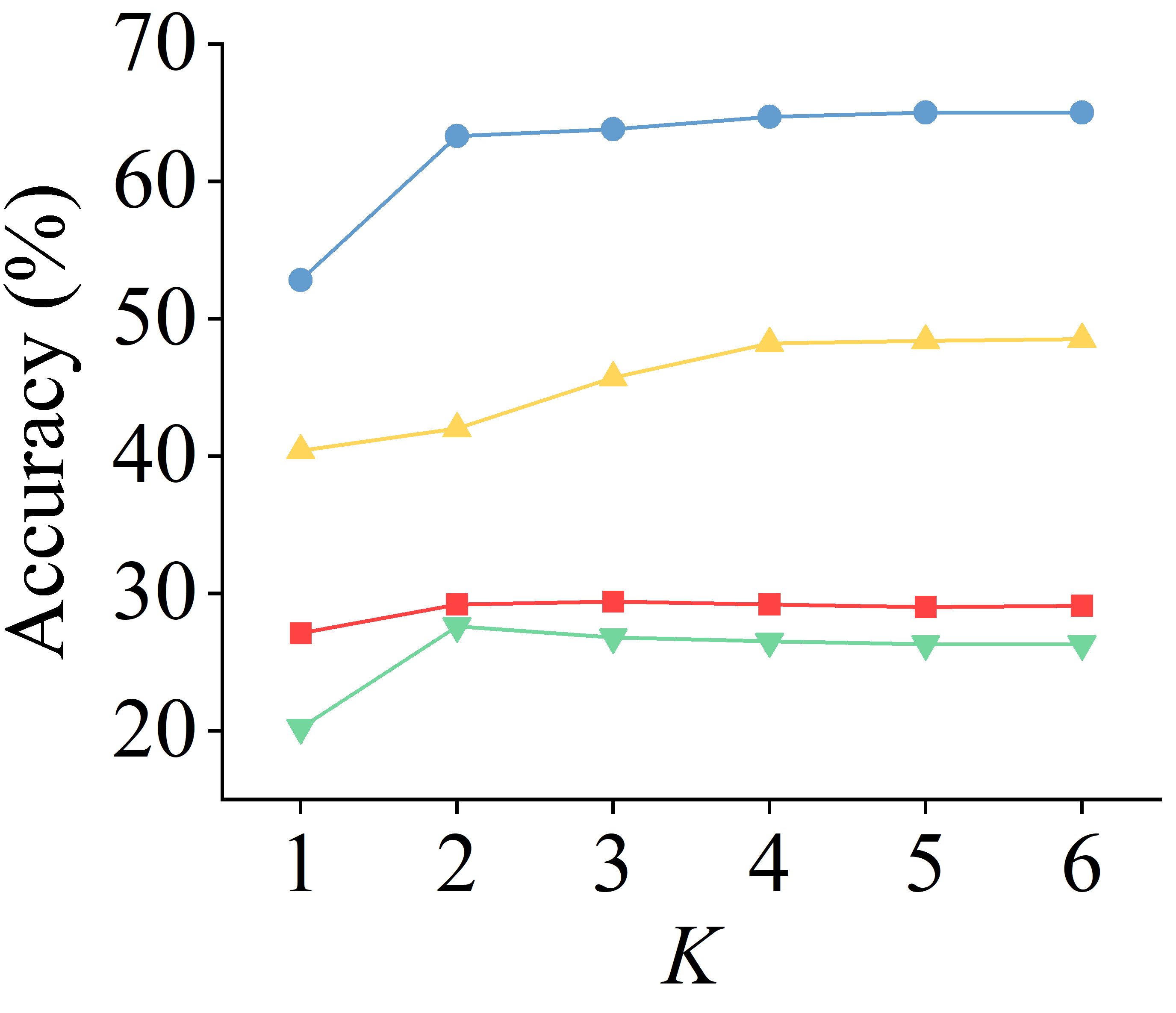}
        \caption{Impact of inference steps in \model.}
        \label{fig.round}
    \end{minipage}
    \hfill
    \begin{minipage}[b]{0.63\linewidth}
        \centering
        \includegraphics[width=\linewidth]{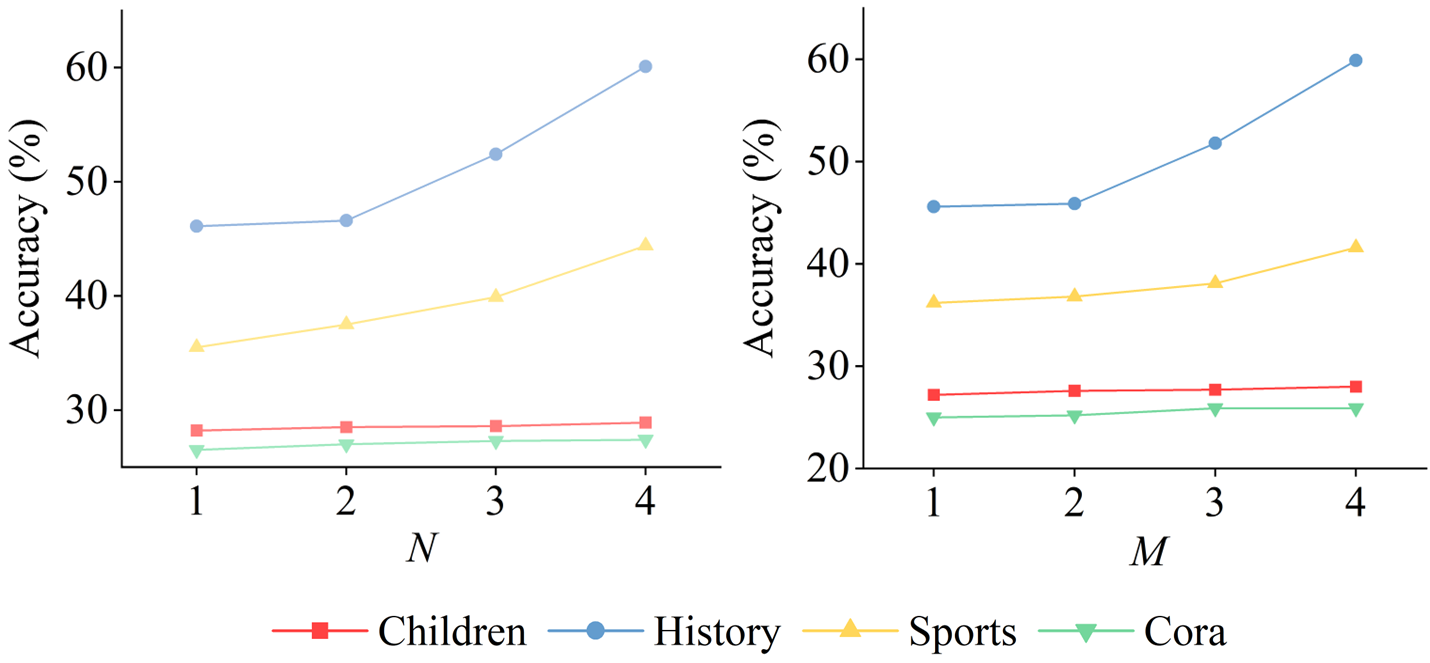}
        \vspace{-3mm}
        \caption{Impact of topological and semantic retrieval size.}
        \label{fig.number}
    \end{minipage}
\end{figure}

\stitle{Effect of inference steps.}
We analyze the impact of the number of reasoning steps $K$ in \model, as shown in Fig.~\ref{fig.round}. When $K=1$, the model reduces to single-step inference without graph-aware retrieval or context refinement, which is equivalent to TEA-GLM. When $K=2$, graph-aware retrieval is introduced in the first step, while context refinement is not yet involved. When $K \geq 3$, the full framework is enabled.
We observe that increasing $K$ from 1 to 2 yields significant improvement, highlighting the effectiveness of graph-aware retrieval. Further increasing $K$ brings additional gains, indicating the benefit of iterative context refinement, although the improvement gradually saturates. In our experiments, we set $K=4$.

\stitle{Effect of retrieval size.}
We analyze the impact of the number of retrieved nodes, including the number of topological neighbors $N$ and semantically similar nodes $M$, as shown in Fig.~\ref{fig.number}. 
We observe that increasing both $N$ and $M$ leads to consistent performance improvements, indicating the benefit of incorporating richer structural and semantic evidence. The gain from enlarging $N$ is generally more pronounced, suggesting that structural context serves as the primary signal, while semantic retrieval provides complementary information. The improvement gradually saturates as $N$ and $M$ increase, implying that most useful information can be captured with a small number of retrieved nodes. Due to the input length limitation of the LLM, we set the maximum number of $N$ and $M$ to 4 in our experiments. In our experiment, we set $N=M=4$.



\section{Conclusion}
In this paper, we proposed \model, a reasoning-acting framework for graph learning. We introduced graph-based retrieval that integrates topological and semantic evidence, along with a context refinement mechanism for multi-step reasoning. By combining retrieval and refinement within a step-by-step process, \model\ constructs informative contexts for prediction. Extensive experiments demonstrate strong performance under zero-shot settings, highlighting the effectiveness of structured reasoning with graph-aware evidence acquisition for graph inference.

\newpage
\bibliographystyle{unsrt}
\bibliography{references}

@article{wang2024llms,
  title={Llms as zero-shot graph learners: Alignment of gnn representations with llm token embeddings},
  author={Wang, Duo and Zuo, Yuan and Li, Fengzhi and Wu, Junjie},
  journal={NeurIPS},
  volume={37},
  pages={5950--5973},
  year={2024}
}

@article{yao2023tree,
  title={Tree of thoughts: Deliberate problem solving with large language models},
  author={Yao, Shunyu and Yu, Dian and Zhao, Jeffrey and Shafran, Izhak and Griffiths, Tom and Cao, Yuan and Narasimhan, Karthik},
  journal={NeurIPS},
  volume={36},
  pages={11809--11822},
  year={2023}
}

@article{wang2022self,
  title={Self-consistency improves chain of thought reasoning in language models},
  author={Wang, Xuezhi and Wei, Jason and Schuurmans, Dale and Le, Quoc and Chi, Ed and Narang, Sharan and Chowdhery, Aakanksha and Zhou, Denny},
  journal={ICLR},
  year={2023}
}

@article{zhou2022least,
  title={Least-to-most prompting enables complex reasoning in large language models},
  author={Zhou, Denny and Sch{\"a}rli, Nathanael and Hou, Le and Wei, Jason and Scales, Nathan and Wang, Xuezhi and Schuurmans, Dale and Cui, Claire and Bousquet, Olivier and Le, Quoc and others},
  journal={ICLR},
  year={2023}
}

@article{chen2024llaga,
  title={Llaga: Large language and graph assistant},
  author={Chen, Runjin and Zhao, Tong and Jaiswal, Ajay and Shah, Neil and Wang, Zhangyang},
  journal={ICML},
  year={2024}
}

@inproceedings{tang2024higpt,
  title={Higpt: Heterogeneous graph language model},
  author={Tang, Jiabin and Yang, Yuhao and Wei, Wei and Shi, Lei and Xia, Long and Yin, Dawei and Huang, Chao},
  booktitle={SIGKDD},
  pages={2842--2853},
  year={2024}
}

@inproceedings{tang2024graphgpt,
  title={Graphgpt: Graph instruction tuning for large language models},
  author={Tang, Jiabin and Yang, Yuhao and Wei, Wei and Shi, Lei and Su, Lixin and Cheng, Suqi and Yin, Dawei and Huang, Chao},
  booktitle={SIGIR},
  pages={491--500},
  year={2024}
}

@article{yuan2025graver,
  title={GRAVER: Generative Graph Vocabularies for Robust Graph Foundation Models Fine-tuning},
  author={Yuan, Haonan and Sun, Qingyun and Shi, Junhua and Fu, Xingcheng and Hooi, Bryan and Li, Jianxin and Yu, Philip S},
  journal={NeurIPS},
  year={2025}
}

@article{wang2024gft,
  title={Gft: Graph foundation model with transferable tree vocabulary},
  author={Wang, Zehong and Zhang, Zheyuan and Chawla, Nitesh V and Zhang, Chuxu and Ye, Yanfang},
  journal={NeurIPS},
  volume={37},
  pages={107403--107443},
  year={2024}
}

@inproceedings{he2025unigraph2,
  title={Unigraph2: Learning a unified embedding space to bind multimodal graphs},
  author={He, Yufei and Sui, Yuan and He, Xiaoxin and Liu, Yue and Sun, Yifei and Hooi, Bryan},
  booktitle={WWW 2025},
  pages={1759--1770},
  year={2025}
}

@inproceedings{wei2022chain,
  title={Chain-of-thought prompting elicits reasoning in large language models},
  author={Wei, Jason and Wang, Xuezhi and Schuurmans, Dale and Bosma, Maarten and Xia, Fei and Chi, Ed and Le, Quoc V and Zhou, Denny and others},
  booktitle={NeurIPS},
  volume={35},
  pages={24824--24837},
  year={2022}
}

@book{cook2006mining,
  title={Mining graph data},
  author={Cook, Diane J and Holder, Lawrence B},
  year={2006},
  publisher={John Wiley \& Sons}
}

@article{xia2021graph,
  title={Graph learning: A survey},
  author={Xia, Feng and Sun, Ke and Yu, Shuo and Aziz, Abdul and Wan, Liangtian and Pan, Shirui and Liu, Huan},
  journal={IEEE Transactions on Artificial Intelligence},
  volume={2},
  number={2},
  pages={109--127},
  year={2021}}

@inproceedings{zhangautomatic,
  title={Automatic Chain of Thought Prompting in Large Language Models},
  author={Zhang, Zhuosheng and Zhang, Aston and Li, Mu and Smola, Alex},
  booktitle={ICLR},
  year={2023}
}

@inproceedings{yao2022react,
  title={React: Synergizing reasoning and acting in language models},
  author={Yao, Shunyu and Zhao, Jeffrey and Yu, Dian and Du, Nan and Shafran, Izhak and Narasimhan, Karthik R and Cao, Yuan},
  booktitle={ICLR},
  year={2022}
}

@inproceedings{feng2023towards,
  title={Towards revealing the mystery behind chain of thought: a theoretical perspective},
  author={Feng, Guhao and Zhang, Bohang and Gu, Yuntian and Ye, Haotian and He, Di and Wang, Liwei},
  journal={NeurIPS},
  volume={36},
  pages={70757--70798},
  year={2023}
}

@inproceedings{you2020graph,
  title={Graph contrastive learning with augmentations},
  author={You, Yuning and Chen, Tianlong and Sui, Yongduo and Chen, Ting and Wang, Zhangyang and Shen, Yang},
  booktitle={NeurIPS},
  volume={33},
  pages={5812--5823},
  year={2020}
}

@inproceedings{liu2023graphprompt,
  title={{GraphPrompt}: Unifying pre-training and downstream tasks for graph neural networks},
  author={Liu, Zemin and Yu, Xingtong and Fang, Yuan and Zhang, Xinming},
  booktitle={WWW},
  pages={417--428},
  year={2023}
}

@inproceedings{velivckovic2018deep,
  title={Deep Graph Infomax},
  author={Veli{\v{c}}kovi{\'c}, Petar and Fedus, William and Hamilton, William L and Li{\`o}, Pietro and Bengio, Yoshua and Hjelm, R Devon},
  booktitle={ICLR},
  year={2018}
}

@incollection{bundy1984breadth,
  title={Breadth-first search},
  author={Bundy, Alan and Wallen, Lincoln},
  booktitle={Catalogue of artificial intelligence tools},
  pages={13--13},
  year={1984}
}

@incollection{taud2017multilayer,
  title={Multilayer perceptron (MLP)},
  author={Taud, Hind and Mas, Jean-Franccois},
  booktitle={Geomatic approaches for modeling land change scenarios},
  pages={451--455},
  year={2017}
}

@article{xia2015learning,
  title={Learning similarity with cosine similarity ensemble},
  author={Xia, Peipei and Zhang, Li and Li, Fanzhang},
  journal={Information sciences},
  volume={307},
  pages={39--52},
  year={2015}}

@inproceedings{kipf2016semi,
  title={Semi-supervised classification with graph convolutional networks},
  author={Kipf, Thomas N and Welling, Max},
  booktitle={ICLR},
  year={2017}
}

@inproceedings{hamilton2017inductive,
  title={Inductive representation learning on large graphs},
  author={Hamilton, Will and Ying, Zhitao and Leskovec, Jure},
  booktitle={NeurIPS},
  year={2017}
}

@inproceedings{fu2025preact,
  title={PreAct: Prediction enhances agent’s planning ability},
  author={Fu, Dayuan and Huang, Jianzhao and Lu, Siyuan and Dong, Guanting and Wang, Yejie and He, Keqing and Xu, Weiran},
  booktitle={ACL},
  pages={1--16},
  year={2025}
}

@inproceedings{shensatori,
  title={Satori: Reinforcement Learning with Chain-of-Action-Thought Enhances LLM Reasoning via Autoregressive Search},
  author={Shen, Maohao and Zeng, Guangtao and Qi, Zhenting and Hong, Zhang-Wei and Chen, Zhenfang and Lu, Wei and Wornell, Gregory W and Das, Subhro and Cox, David Daniel and Gan, Chuang},
  booktitle={ICML},
  year={2025}
}

@inproceedings{velivckovic2017graph,
  title={Graph attention networks},
  author={Veli{\v{c}}kovi{\'c}, Petar and Cucurull, Guillem and Casanova, Arantxa and Romero, Adriana and Lio, Pietro and Bengio, Yoshua},
  booktitle={ICLR},
  year={2018}
}

@inproceedings{xu2018powerful,
  title={How powerful are graph neural networks?},
  author={Xu, Keyulu and Hu, Weihua and Leskovec, Jure and Jegelka, Stefanie},
  booktitle={ICLR},
  year={2019}
}

@inproceedings{wen2023augmenting,
  title={Augmenting Low-Resource Text Classification with Graph-Grounded Pre-training and Prompting},
  author={Wen, Zhihao and Fang, Yuan},
  booktitle={SIGIR},
  year={2023}
}

@inproceedings{ma2021homophily,
  title={Is homophily a necessity for graph neural networks?},
  author={Ma, Yao and Liu, Xiaorui and Shah, Neil and Tang, Jiliang},
  booktitle={ICLR},
  year={2022}
}

@article{luan2022revisiting,
  title={Revisiting heterophily for graph neural networks},
  author={Luan, Sitao and Hua, Chenqing and Lu, Qincheng and Zhu, Jiaqi and Zhao, Mingde and Zhang, Shuyuan and Chang, Xiao-Wen and Precup, Doina},
  journal={NeurIPS},
  pages={1362--1375},
  year={2022}
}

@inproceedings{liu2023one,
  title={One for All: Towards Training One Graph Model for All Classification Tasks},
  author={Liu, Hao and Feng, Jiarui and Kong, Lecheng and Liang, Ningyue and Tao, Dacheng and Chen, Yixin and Zhang, Muhan},
  booktitle={ICLR},
  year={2024}
}

@article{yu2023generalized,
  title={Generalized Graph Prompt: Toward a Unification of Pre-Training and Downstream Tasks on Graphs},
  author={Yu, Xingtong and Liu, Zhenghao and Fang, Yuan and Liu, Zemin and Chen, Sihong and Zhang, Xinming},
  journal={IEEE TKDE},
  volume={36}, 
  number={11},
  pages={6237- 6250},
  year={2023}
}

@inproceedings{hu2020open,
  title={Open graph benchmark: Datasets for machine learning on graphs},
  author={Hu, Weihua and Fey, Matthias and Zitnik, Marinka and Dong, Yuxiao and Ren, Hongyu and Liu, Bowen and Catasta, Michele and Leskovec, Jure},
  booktitle={NeurIPS},
  year={2020}
}

@article{yu2025gcot,
  title={{GCoT}: Chain-of-Thought Prompt Learning for Graphs},
  author={Yu, Xingtong and Zhou, Chang and Kuai, Zhongwei and Zhang, Xinming and Fang, Yuan},
  journal={arXiv preprint arXiv:2502.08092},
  year={2025}
}

@article{yang2022geometric,
  title={Geometric knowledge distillation: Topology compression for graph neural networks},
  author={Yang, Chenxiao and Wu, Qitian and Yan, Junchi},
  journal={NeurIPS},
  volume={35},
  pages={29761--29775},
  year={2022}
}

@inproceedings{zhang2022graphless,
  title={Graph-less neural networks: Teaching old MLPs new tricks via distillation},
  author={Zhang, Shichang and Liu, Yozen and Sun, Yizhou and Shah, Neil},
  booktitle={ICLR},
  year={2022}
}

@article{wu2022nodeformer,
  title={Nodeformer: A scalable graph structure learning transformer for node classification},
  author={Wu, Qitian and Zhao, Wentao and Li, Zenan and Wipf, David P and Yan, Junchi},
  journal={NeurIPS},
  volume={35},
  pages={27387--27401},
  year={2022}
}

@inproceedings{wu2023difformer,
  title={{DIFFormer}: Scalable (graph) transformers induced by energy constrained diffusion},
  author={Wu, Qitian and Yang, Chenxiao and Zhao, Wentao and He, Yixuan and Wipf, David and Yan, Junchi},
  booktitle={International Conference on Learning Representations},
  year={2023}
}

@article{chiang2023vicuna,
  title={Vicuna: An open-source chatbot impressing gpt-4 with 90\%* chatgpt quality},
  author={Chiang, Wei-Lin and Li, Zhuohan and Lin, Ziqing and Sheng, Ying and Wu, Zhanghao and Zhang, Hao and Zheng, Lianmin and Zhuang, Siyuan and Zhuang, Yonghao and Gonzalez, Joseph E and others},
  journal={See https://vicuna. lmsys. org (accessed 14 April 2023)},
  volume={2},
  number={3},
  pages={6},
  year={2023}
}

@inproceedings{he2024harnessing,
  title={Harnessing explanations: LLM-to-LM interpreter for enhanced text-attributed graph representation learning},
  author={He, Xiaoxin and Bresson, Xavier and Laurent, Thomas and Perold, Adam and LeCun, Yann and Hooi, Bryan},
  booktitle={ICLR},
  year={2024}
}

@article{yan2023comprehensive,
  title={A comprehensive study on text-attributed graphs: Benchmarking and rethinking},
  author={Yan, Hao and Li, Chaozhuo and Long, Ruosong and Yan, Chao and Zhao, Jianan and Zhuang, Wenwen and Yin, Jun and Zhang, Peiyan and Han, Weihao and Sun, Hao and others},
  journal={NeurIPS},
  volume={36},
  pages={17238--17264},
  year={2023}
}

\newpage
\appendix
\section{Alogrithm}
We summarize the overall procedure of \model\ in Algorithm~\ref{alg:model}. 
Lines 2--4 initialize the node representation by encoding the target node and projecting it into the LLM embedding space, followed by constructing the initial instruction. 
In lines 6--8, the model performs the first reasoning step and invokes graph-based retrieval to obtain both topological and semantic summaries, which are used to build the initial context. 
Then, lines 10--13 implement multi-step reasoning via context refinement, where at each step the LLM generates a new thought based on the current context, and the context is updated through a refinement action that distills and reorganizes the accumulated information. 
Finally, in lines 14--15, the model produces the prediction based on the refined context after $K$ steps.

\section{Further Descriptions of Datasets}
\label{app:dataset_description}

We summarize all datasets in Table~\ref{tab:dataset} and provide further comprehensive descriptions of these datasets.

\begin{itemize}
    \item \textbf{Arxiv}~\cite{hu2020open} is a large-scale citation network built from Computer Science papers on the arXiv preprint server. The graph contains 169{,}343 paper nodes and 1{,}166{,}243 citation edges, with labels covering 40 arXiv CS sub-categories.

    \item \textbf{PubMed}~\cite{he2024harnessing} consists of 19{,}717 diabetes-related publications connected by 44{,}338 citation links. Each node is labeled as one of three categories: experimentally induced diabetes, type 1 diabetes, or type 2 diabetes.

    \item \textbf{Cora}~\cite{wen2023augmenting}, formally known as the Cora Research Paper Classification Dataset, is an expanded citation graph for research paper classification. It includes 25{,}120 papers and 91{,}140 citation edges, where the 70 node labels correspond to fine-grained research categories.

    \item \textbf{Computer}~\cite{yan2023comprehensive} is an e-commerce graph from the TAG benchmark, extracted from computer-related products in Amazon-Electronics. It has 87{,}229 product nodes, 721{,}081 co-viewing or co-purchasing edges, and 10 third-level product categories.

    \item \textbf{Photo}~\cite{yan2023comprehensive} is constructed from photo-related products in Amazon-Electronics. The graph contains 48{,}362 products and 500{,}928 behavioral links, where an edge indicates that two products are frequently co-viewed or co-purchased. The prediction labels are 12 third-level product categories.

    \item \textbf{Children}~\cite{yan2023comprehensive} is an Amazon-Books graph whose nodes are children's book products. It comprises 76{,}875 nodes and 1{,}554{,}578 co-viewing or co-purchasing edges, with 24 labels defined by third-level book categories.

    \item \textbf{History}~\cite{yan2023comprehensive} is another Amazon-Books graph, focusing on history-related books. It contains 41{,}551 book nodes, 358{,}574 product-relation edges, and 12 third-level category labels.

    \item \textbf{Sports}~\cite{yan2023comprehensive} is an Amazon-Sports graph built from fitness-related products. It is the largest e-commerce target dataset in our experiments, with 173{,}055 product nodes, 1{,}773{,}500 co-viewing or co-purchasing edges, and 13 fine-grained product categories.
\end{itemize}

\begin{algorithm}[t]
\small
\caption{\textsc{Graph-aware Reasoning and Acting}}
\label{alg:model}
\begin{algorithmic}[1]
\Require Graph $G=(\bV,\bE)$, target node $v$, node text $x^{\mathrm{text}}$, 
projection $\mathtt{Proj}(\cdot;\phi)$, frozen LLM, steps $K$
\Ensure Prediction $\hat{y}$

\State \textbf{Initialization:}
\State $\mathbf{h}_v \leftarrow \mathtt{GraphEncoder}(G, v)$
\State $\mathbf{h}_v^{\mathrm{tok}} \leftarrow \mathtt{Proj}(\mathbf{h}_v;\phi)$
\State Construct instruction $\mathcal{I}$ from $(x^{\mathrm{text}}, \mathbf{h}_v^{\mathrm{tok}})$

\State \textbf{Graph-based retrieval:}
\State $\text{Thought}^1 \leftarrow \mathtt{LLM}(\mathcal{I}, \mathcal{Q})$
\State $\{s^{\mathrm{top}}, s^{\mathrm{sem}}\} \leftarrow \mathtt{Act}_{\text{retrieve}}(G, v)$
\State $\mathcal{C}^1 \leftarrow \mathtt{Init}(\text{Thought}^1,\; \{s^{\mathrm{top}}, s^{\mathrm{sem}}\})$

\State \textbf{Context refinement:}
\For{$k = 2$ to $K$}
    \State $\text{Thought}^{k+1} \leftarrow \mathtt{LLM}(\mathcal{I},\; \mathcal{C}^k,\; \mathcal{Q})$
    \State $\text{Observation}^{k+1} \leftarrow \mathtt{Act}_{\text{refine}}(\mathcal{C}^k,\; \text{Thought}^{k+1})$
    \State $\mathcal{C}^{k+1} \leftarrow \mathtt{Update}(\mathcal{C}^k,\; \text{Thought}^{k+1},\; \text{Observation}^{k+1})$
\EndFor
\State \textbf{Final prediction:}
\State $\hat{y} \leftarrow \mathtt{LLM}(\mathcal{I},\; \mathcal{C}^K,\; \mathcal{Q})$
\State \Return $\hat{y}$
\end{algorithmic}
\end{algorithm}

\section{Further Descriptions of Baselines}
\label{app:baseline_description}

In this section, we provide additional details about the baselines used in our experiments.

\stitle{(1) Non-graph model.}
\begin{itemize}
    \item \textbf{MLP}~\citep{taud2017multilayer}: A multilayer perceptron that predicts node labels based only on node features. It does not explicitly model graph topology, and therefore serves as a structure-agnostic baseline for evaluating whether graph structural information is necessary.
\end{itemize}

\stitle{(2) Supervised graph methods.}
\begin{itemize}
    \item \textbf{GCN}~\citep{kipf2016semi}: A representative graph neural network that aggregates and transforms information from local neighborhoods through graph convolution. It captures structural information by recursively propagating node features over the graph.

    \item \textbf{GraphSAGE}~\citep{hamilton2017inductive}: An inductive graph representation learning method that samples and aggregates neighborhood information. It learns node representations by combining target node features with sampled neighbor representations, making it suitable for large-scale graphs.

    \item \textbf{GAT}~\citep{velivckovic2017graph}: A graph attention network that introduces attention weights into neighborhood aggregation. Instead of treating all neighbors equally, GAT assigns different importance scores to neighboring nodes, allowing the model to selectively aggregate more informative structural evidence.

    \item \textbf{NodeFormer}~\citep{wu2022nodeformer}: A scalable graph Transformer for node classification. It models node interactions through an efficient attention mechanism and is designed to handle large-scale graph structure learning.

    \item \textbf{DIFFormer}~\citep{wu2023difformer}: A scalable graph Transformer induced by energy-constrained diffusion. It captures long-range dependencies on graphs through a diffusion-based formulation, improving the ability to model global structural information.
\end{itemize}

\stitle{(3) Self-supervised graph methods.}
\begin{itemize}
    \item \textbf{DGI}~\citep{velivckovic2018deep}: A self-supervised graph representation learning method based on mutual information maximization. It learns node embeddings by contrasting local node representations with a global graph summary, without relying on task-specific labels during representation learning.
\end{itemize}

\stitle{(4) Graph knowledge distillation.}
\begin{itemize}
    \item \textbf{GKD}~\citep{yang2022geometric}: A graph knowledge distillation framework that transfers structural knowledge from a teacher GNN trained on complete graph information to a student model. It aims to preserve useful topological knowledge while reducing the dependency on full graph access.

    \item \textbf{GLNN}~\citep{zhang2022graphless}: A graph-less neural network framework that distills graph-aware knowledge from GNNs into an MLP-like architecture. By transferring graph-aware predictions into a structure-free student model, GLNN reduces the reliance on graph connectivity during inference.
\end{itemize}

\stitle{(5) Large language models.}
\begin{itemize}
    \item \textbf{Vicuna-7B-v1.5}~\citep{chiang2023vicuna}: An instruction-tuned open-source large language model. In our experiments, Vicuna directly performs prediction based on textual node information, without explicitly using graph structure.

    \item \textbf{Vicuna-7B-SPT}: A soft prompt tuning variant of Vicuna-7B-v1.5. It introduces learnable soft prompts to adapt the LLM to graph-related prediction tasks while still relying mainly on textual information.
\end{itemize}

\stitle{(6) Graph--LLM methods.}
\begin{itemize}
    \item \textbf{OFA}~\citep{liu2023one}: A unified graph learning framework designed to handle different graph classification tasks with a shared model. It improves transferability across tasks by formulating graph learning problems under a unified task interface.

    \item \textbf{GraphGPT}~\citep{tang2024graphgpt}: A graph instruction tuning framework that aligns graph representations with large language models. We report two variants following the TEA-GLM setting: \textbf{GraphGPT-std}, which uses the standard graph instruction tuning pipeline, and \textbf{GraphGPT-cot}, which further incorporates Chain-of-Thought-style instruction data generated by LLMs.

    \item \textbf{LLaGA}~\citep{chen2024llaga}: A graph--language assistant that connects graph structural information with LLMs for graph-related tasks. It uses graph-aware representations to support LLM-based prediction and reasoning.

    \item \textbf{TEA-GLM}~\citep{wang2024llms}: A token embedding-aligned graph language model that aligns GNN representations with the token embedding space of a frozen LLM. It maps graph representations into a fixed number of graph token embeddings and inserts them into instruction templates for zero-shot graph learning. Since \model\ is built upon the TEA-GLM graph--LLM interface, TEA-GLM is used as the most direct backbone baseline for evaluating the effectiveness of our graph-aware reasoning-acting mechanism.
\end{itemize}

\begin{table}
\centering
\caption{Summary of datasets.}
\label{tab:dataset}
\begin{tabular}{c c c c c}
\toprule
\multicolumn{1}{c}{Domain} & \multicolumn{1}{c}{Dataset} & \multicolumn{1}{c}{\#Nodes} & \multicolumn{1}{c}{\#Edges} & \multicolumn{1}{c}{\#Classes} \\
\midrule
\multirow{3}{*}{Citation}  & Arxiv      & 169,343 & 1,166,243 & 40 \\
                           & Pubmed     & 19,717  & 44,338    & 3  \\
                           & Cora       & 25,120  & 91,140    & 70 \\
\midrule
\multirow{5}{*}{E-commerce} & Ele-Computer & 87,229  & 721,081   & 10 \\
                            & Ele-Photo    & 48,362  & 500,928   & 12 \\
                            & Book-Children & 76,875  & 1,554,578 & 24 \\
                            & Book-History & 41,551  & 358,574   & 12 \\
                            & Sports-Fitness & 173,055 & 1,773,500 & 13 \\
\bottomrule
\end{tabular}
\end{table}

\section{Implementation Details}
\label{app:implementation_details}

\stitle{Environment.}
All experiments are conducted on Ubuntu 22.04 with a 25 vCPU Intel(R) Xeon(R) Platinum 8481C and an vGPU-48GB with 48GB memory.

\stitle{Optimizer.}
For all experiment, Adam is used as the optimizer.

\stitle{Details of baselines.}
For fair and consistent comparison, we directly adopt the reported results of all non-TEA-GLM baseline methods from TEA-GLM~\cite{wang2024llms}, including MLP, GCN, \method{GraphSage}, GAT, DGI, GKD, GLNN, \method{NodeFormer}, DIFFormer,  \method{Vicuna-7B-v1.5}, \method{Vicuna-7B-SPT}, \method{GraphGPT}, \method{LLaGA}, OFA and TEA-GLM. 
These baselines are evaluated under the same dataset setting and evaluation metrics as TEA-GLM. 

For TEA-GLM, we follow its recommended setting: raw node texts are encoded by a pretrained BERT model, GraphSAGE is used as the graph encoder, Vicuna-7B-v1.5 is used as the frozen LLM backbone, and the number of GNN layers is set to 2.
The graph encoder is pretrained on the source dataset, and the linear projector maps graph representations into graph token embeddings for zero-shot inference on unseen target datasets.

\stitle{Implementation details of \model.}
We adopt a 3-layer GraphSAGE encoder as the graph backbone to obtain node representations, which are then mapped into the LLM embedding space via a linear projection layer. 
For graph-aware acting, we retrieve a fixed number of nodes from both structural and semantic perspectives. Specifically, we select $N=4$ neighbors for topological retrieval using breadth-first traversal, and $M=4$ nodes for semantic retrieval based on embedding similarity.
For multi-step reasoning, we set the number of inference steps to $K=4$.
The prompt design used in our experiments is provided in Sect.~\ref{app.sec.prompt}.

\section{Supervised Performance}
\begin{table}[H]
    \centering
    \caption{Accuracy and Macro-F1 on training datasets.}
    \label{tab:supervised}
    \setlength{\tabcolsep}{4pt}
    \resizebox{0.8\linewidth}{!}{
    \begin{tabular}{l|cc|cc}
    \toprule
    \multirow{2}{*}{\textbf{Model}} & \multicolumn{2}{c|}{\textbf{Arxiv}} & \multicolumn{2}{c}{\textbf{Computer}} \\
    \cmidrule{2-5}
    & Acc & F1 & Acc & F1 \\
    \midrule
    \method{MLP} & 0.546$\pm$0.004 & 0.295$\pm$0.007 & 0.420$\pm$0.006 & 0.267$\pm$0.005  \\
    \midrule
    \method{GCN} & 0.545$\pm$0.005 & 0.317$\pm$0.006 & 0.424$\pm$0.012 & 0.386$\pm$0.014  \\
    \method{GraphSAGE} & 0.556$\pm$0.006 & 0.315$\pm$0.008 & 0.534$\pm$0.037 & 0.347$\pm$0.036  \\
    \method{GAT} & 0.561$\pm$0.003 & 0.339$\pm$0.005 & 0.609$\pm$0.035 & \underline{0.598}$\pm$0.039  \\
    \method{NodeFormer} & 0.544$\pm$0.016 & 0.297$\pm$0.029 & 0.434$\pm$0.012 & 0.288$\pm$0.012  \\
    \method{DIFFormer} & 0.616$\pm$0.025 & 0.356$\pm$0.024 & 0.629$\pm$0.012 & 0.467$\pm$0.022  \\
    \midrule
    \method{DGI} & 0.342$\pm$0.024 & 0.336$\pm$0.011 & 0.594$\pm$0.004 & 0.452$\pm$0.008  \\
    \midrule
    \method{GKD} & 0.393$\pm$0.085 & 0.164$\pm$0.029 & 0.351$\pm$0.031 & 0.155$\pm$0.016  \\
    \method{GLNN} & 0.602$\pm$0.004 & 0.362$\pm$0.008 & 0.393$\pm$0.005 & 0.243$\pm$0.007  \\
    \midrule
    \method{Vicuna-7B-v1.5} & 0.347$\pm$0.000 & 0.164$\pm$0.001 & 0.372$\pm$0.010 & 0.304$\pm$0.002 \\
    \midrule
    \method{OFA} & 0.682$\pm$0.006 & 0.495$\pm$0.006 & \textbf{0.753}$\pm$0.004 & \textbf{0.687}$\pm$0.006  \\
    \method{GraphGPT-std} & 0.626 & 0.262 & - & - \\
    \method{GraphGPT-cot} & 0.576 & 0.228 & - & - \\
    \method{LLaGA} & \textbf{0.749}$\pm$0.001 & \underline{0.575}$\pm$0.003 & 0.642$\pm$0.004 & 0.562$\pm$0.001 \\
    \method{TEA-GLM} & 0.655$\pm$0.001 & 0.445$\pm$0.002 & 0.578$\pm$0.002 & 0.496$\pm$0.010 \\
    \method{\model} & \underline{0.738}$\pm$0.011 & \textbf{0.645}$\pm$0.012 & \underline{0.687}$\pm$0.007 & 0.513$\pm$0.015 \\
    \bottomrule
    \end{tabular}
    }

    \vspace{1mm}
\end{table}
We further evaluate \model\ under a supervised setting, where models are trained on the Arxiv and Computer datasets and evaluated on held-out test splits. The results are reported in Table~\ref{tab:supervised}.
We make the following observations.
First, \model\ achieves competitive performance compared with existing baselines, while not consistently outperforming state-of-the-art graph--LLM methods in this setting. This is expected, as supervised methods can directly optimize task-specific objectives with labeled data, whereas \model\ is primarily designed for reasoning-based inference rather than end-to-end supervised training.
Second, \model\ demonstrates relatively stronger performance in terms of Macro-F1, particularly on Arxiv, indicating improved class-level balance. This suggests that the reasoning-acting mechanism, which dynamically aggregates and refines evidence, can help mitigate bias toward dominant classes even in supervised scenarios.
Noote that the supervised setting is not the primary focus of this work. Instead, \model\ is designed to address the more challenging zero-shot graph learning scenario, where labeled data is unavailable and effective reasoning over structured context becomes critical. As demonstrated in Sect.~\ref{sec.performance}, \model\ achieves significant improvements under zero-shot settings, highlighting the advantage of reasoning-acting synergy for cross-dataset generalization.

\section{Macro-F1 Results}
\begin{table}
  \centering
  \caption{Macro-F1 score of node classification.}
  \label{tab:macrof1}
  \setlength{\tabcolsep}{4pt}
  \resizebox{\linewidth}{!}{
  \begin{tabular}{lcc|cccc}
    \toprule
    Model & \multicolumn{2}{c|}{Citation} & \multicolumn{4}{c}{E-commerce} \\
    \cmidrule(r){2-3} \cmidrule(l){4-7}
    & Cora & Pubmed & Children & History & Photo & Sports \\
    \midrule
    \method{MLP} & 0.009$\pm$0.004 & 0.246$\pm$0.042 & 0.007$\pm$0.007 & 0.023$\pm$0.008 & 0.041$\pm$0.023 & 0.019$\pm$0.005 \\
    \midrule
    \method{GCN}          & 0.007$\pm$0.001 & 0.187$\pm$0.021 & 0.006$\pm$0.004 & 0.024$\pm$0.013 & 0.034$\pm$0.007 & 0.017$\pm$0.009 \\
    \method{GraphSAGE}    & 0.007$\pm$0.003 & 0.257$\pm$0.084 & 0.005$\pm$0.003 & 0.029$\pm$0.024 & 0.020$\pm$0.011 & 0.021$\pm$0.004 \\
    \method{GAT}          & 0.006$\pm$0.001 & 0.259$\pm$0.065 & 0.063$\pm$0.067 & 0.159$\pm$0.117 & 0.036$\pm$0.035 & 0.091$\pm$0.090 \\
    \method{NodeFormer}   & 0.008$\pm$0.003 & 0.232$\pm$0.089 & 0.019$\pm$0.008 & 0.046$\pm$0.031 & 0.055$\pm$0.006 & 0.049$\pm$0.009 \\
    \method{DIFFormer}    & 0.007$\pm$0.002 & 0.187$\pm$0.007 & 0.002$\pm$0.002 & 0.050$\pm$0.019 & 0.069$\pm$0.010 & 0.045$\pm$0.007 \\
    \midrule
    \method{DGI}          & 0.004$\pm$0.002 & 0.213$\pm$0.127 & 0.012$\pm$0.004 & 0.038$\pm$0.015 & 0.045$\pm$0.015 & 0.018$\pm$0.005 \\
    \midrule
    \method{GKD}          & 0.004$\pm$0.001 & 0.247$\pm$0.039 & 0.028$\pm$0.003 & 0.060$\pm$0.008 & 0.049$\pm$0.015 & 0.050$\pm$0.008 \\
    \method{GLNN}         & 0.006$\pm$0.001 & 0.221$\pm$0.033 & 0.021$\pm$0.003 & 0.064$\pm$0.007 & 0.057$\pm$0.002 & 0.052$\pm$0.003 \\
    \midrule
    \method{Vicuna-7B-v1.5}  & 0.109$\pm$0.002 & 0.629$\pm$0.024 & \textbf{0.279}$\pm$0.002 & \underline{0.349}$\pm$0.003 & 0.383$\pm$0.001 & 0.410$\pm$0.002 \\
    \midrule
    \method{OFA}          & 0.091$\pm$0.013 & 0.287$\pm$0.059 & 0.017$\pm$0.010 & 0.026$\pm$0.007 & 0.103$\pm$0.007 & 0.043$\pm$0.021 \\
    \method{GraphGPT-std} & 0.082 & 0.649 & --- & --- & --- & --- \\
    \method{GraphGPT-cot} & \textbf{0.127} & 0.482 & --- & --- & --- & --- \\
    \method{LLaGA}        & 0.108$\pm$0.014 & \underline{0.778}$\pm$0.056 & 0.163$\pm$0.029 & 0.144$\pm$0.025 & 0.362$\pm$0.039 & \underline{0.446}$\pm$0.035 \\
    \method{TEA-GLM}      & 0.107$\pm$0.012 & \textbf{0.846}$\pm$0.011 & 0.209$\pm$0.010 & 0.336$\pm$0.021 & \underline{0.404}$\pm$0.017 & 0.396$\pm$0.007 \\
    \model     & \underline{0.116}$\pm$0.016 & 0.759$\pm$0.017 & \underline{0.216}$\pm$0.008 & \textbf{0.351}$\pm$0.022 & \textbf{0.409}$\pm$0.011 & \textbf{0.464}$\pm$0.009 \\
    \bottomrule
  \end{tabular}
  }

  \vspace{1mm}
  \parbox{\linewidth}{\scriptsize The best method is bolded and the runner-up is underlined.}
\end{table}
We further report Macro-F1 scores for zero-shot node classification in Table~\ref{tab:macrof1}. 
Overall, \model\ achieves competitive or superior performance on most datasets, particularly in the e-commerce domain, where it consistently outperforms existing graph--LLM methods. 
Compared with accuracy, the improvements in Macro-F1 are more pronounced, indicating that \model\ provides better class-level balance under zero-shot settings. 
This can be attributed to the reasoning-acting mechanism, which dynamically aggregates and refines evidence from both structural and semantic sources, thereby reducing bias toward dominant classes.
On citation datasets, \model\ remains competitive with strong baselines such as TEA-GLM, showing that the proposed framework maintains robust performance across different domains.

\begin{figure}[t]
    \centering
    \includegraphics[width=0.3\linewidth]{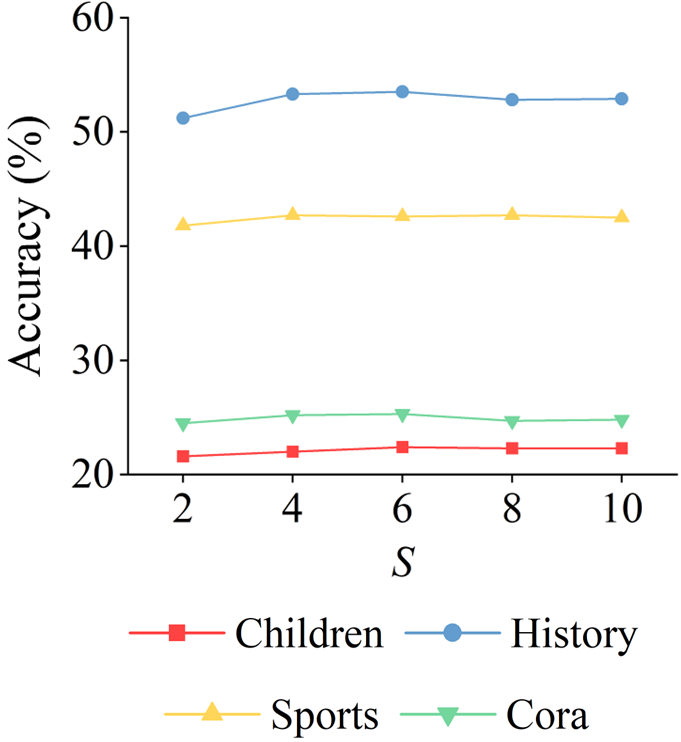}
    \vspace{-2mm}
    \caption{Numbers of Wiki entities for Search.}
    \label{fig.search}
\end{figure}
\section{Effect of the Number of Retrieved Entities in Textual Search}\label{app.sec.search}
We further analyze the impact of the number of retrieved Wikipedia entities $S$ in the text-based Search action, as shown in Fig.~\ref{fig.search}. We observe that increasing $S$ brings only marginal improvements across all datasets. The performance slightly improves when increasing $S$ from 2 to 4, indicating that a small amount of external evidence can provide limited complementary information. However, further increasing $S$ does not lead to consistent gains and quickly reaches saturation, with performance remaining largely stable thereafter. This suggests that simply retrieving more external textual entities does not substantially enhance graph reasoning, and may introduce redundant or noisy information. The results further confirm that external search plays a limited role compared to graph-aware evidence.

\section{Prompt Design} \label{app.sec.prompt}
\begin{longtable}{@{}>{\centering\arraybackslash}p{2cm}@{}p{12cm}@{}}
\caption{Prompt design for node classification on children.} \\
\toprule
Type & Prompts \\
\midrule
\endfirsthead
\toprule
Type & Prompts \\
\midrule
\endhead
\bottomrule
\endfoot
Task Instruction  & Given a representation of a book: <Token 1>, with the following information:\\
    & Description: \{description\}\\
    & Title: \{title\}\\
    & Question: Which category does this book belong to? Please directly give the most likely answer from the following categories: \{categories\} \\
\midrule
Thought Generation  & USER: The task is to classify a children's book by its main topic and content type. Your goal is to identify what kind of story or subject the book presents to young readers (for example, adventure, humor, fantasy, education, or everyday life). Based on the book information above, generate one concise sentence as your current thought to help solve the task.\\
    & ASSISTANT: \\
\midrule
Topological retrieval & USER: You are given a subgraph consisting of a target node and its neighboring nodes from a children's book e-commerce graph. Each neighbor is described by a title and a description. Your task is to infer the most likely category of the subgraph based ONLY on the textual information of the neighbors, and generate a one-sentence summary in the following fixed format: "This subgraph xxx, so the category might be xxx."\\
    & \\
    & IMPORTANT RULES (must be followed strictly):\\
    & 1. Text-based inference: Analyze the titles and descriptions of all neighbors as a whole. Identify the dominant themes, genres, or target age group, such as fantasy, animals, educational concepts, activities, or everyday life experiences. Give more weight to recurring elements shared across neighbors, like magical creatures, vehicle types, specific emotions, or learning topics.\\
    & 2. Noise filtering: Ignore non-informative promotional content, including but not limited to: bestseller status, author biography details (e.g., "has written over one hundred books"), series accolades, review quotes, and generic publisher blurbs. Focus only on information that reflects the book's core story, subject matter, characters, or intended educational/entertainment purpose for children.\\
    & 3. Category constraint: The final category MUST be chosen from the following list and cannot be invented or modified: \{categories\}\\
    & 4. Output constraint: Output exactly ONE sentence. Use the fixed format exactly as specified. Do not add explanations, bullet points, or extra text. The phrase "This subgraph xxx" should concisely describe the overall thematic focus of the children's books in the subgraph.\\
    & \\
    & Here are correct examples:\\
    & - "This subgraph features stories about fairies and magical adventures, so the category might be Fairy Tales, Folk Tales \& Myths."\\
    & - "This subgraph contains books about trucks, trains, and airplanes, so the category might be Cars, Trains \& Things That Go."\\
    & - "This subgraph explores concepts like feelings, friendship, and school life, so the category might be Growing Up \& Facts of Life."\\
    & \\
    & Here is the neighbor information: \{nodes\}\\
    & Please give your reply: \\
\midrule
Semantic retrieval & USER: You are given a center node and some similar nodes, which are based on similar semantic node embeddings, from a children's book e-commerce graph. Each node is described by a title and a description. Your task is to infer the most likely category of the node set based ONLY on the textual information, and generate a one-sentence summary in the following fixed format: "This node set xxx, so the category might be xxx."\\
    & \\
    & IMPORTANT RULES (must be followed strictly):\\
    & 1. Text-based inference: Analyze the titles and descriptions of all nodes as a whole. Identify the dominant themes, genres, or target age group, such as fantasy, animals, educational concepts, activities, or everyday life experiences. Give more weight to recurring elements shared across the node set, like magical creatures, vehicle types, specific emotions, or learning topics.\\
    & 2. Noise filtering: Ignore non-informative promotional content, including but not limited to: bestseller status, author biography details (e.g., "has written over one hundred books"), series accolades, review quotes, and generic publisher blurbs. Focus only on information that reflects the book's core story, subject matter, characters, or intended educational/entertainment purpose for children.\\
    & 3. Category constraint: The final category MUST be chosen from the following list and cannot be invented or modified: \{categories\}\\
    & 4. Output constraint: Output exactly ONE sentence. Use the fixed format exactly as specified. Do not add explanations, bullet points, or extra text. The phrase "This node set xxx" should concisely describe the overall thematic focus of the children's books in the set.\\
    & \\
    & Here are correct examples:\\
    & - "This node set features stories about fairies and magical adventures, so the category might be Fairy Tales, Folk Tales \& Myths."\\
    & - "This node set contains books about trucks, trains, and airplanes, so the category might be Cars, Trains \& Things That Go."\\
    & - "This node set explores concepts like feelings, friendship, and school life, so the category might be Growing Up \& Facts of Life."\\
    & \\
    & Here is the node information: \{nodes\}\\
    & Please give your reply: \\
\midrule
Context refinement & USER: Please summarize the key points from our discussion so far in one sentence. Focus on the most important information that will help answer the original question.\\
    & ASSISTANT: \\
\midrule
Textual Search & USER: To better understand the children's book above, select ONE specific story theme, character type, literary genre, or educational concept that is central to the book and can be searched directly on Wikipedia. Avoid overly generic terms like 'story' or 'children'. Output only a short phrase (maximum 5 words). Do NOT explain.\\
    & ASSISTANT:\\
\midrule
Full prompt example  & A chat between a curious user and an artificial intelligence assistant. The assistant gives helpful, detailed, and polite answers step by step to the user's questions. USER: Given a representation of a book: <Token 1> <Token 2> <Token 3> <Token 4> <Token 5>, with the following information:\\
    & Description: Daisy Meadows has written over one hundred books for children. Her RAINBOW MAGIC series is a New York Times bestseller!\\
    & Title: Amber: The Orange Fairy (Rainbow Magic: The Rainbow Fairies, No. 2).\\
    & \\
    & Thought: The book is a fantasy story featuring a fairy character (Amber the Orange Fairy) as part of a magical rainbow-themed series aimed at young children.\\
    & \\
    & In addition to the information above, the book is part of an e-commerce graph, where neighboring nodes represent children's books that are often related or similar. As a result, adjacent books may share the same or closely related categories. The following summary describes the content and category tendencies of the neighboring books, and can be used as contextual evidence to help answer the question.\\
    & Neighbor summary: This subgraph features books written by Daisy Meadows, with a focus on the RAINBOW MAGIC series, which is a New York Times bestseller. The category might be Literature \& Fiction or Children's Books.\\
    & \\
    & Here is another summary generated from nodes that are similar to the center node.\\
    & Node summary: This node set features humorous stories and books about everyday life experiences, so the category might be Humor.\\
    & \\
    & Thought: The book might belong in the Children's Books category with a focus on fantasy and humorous stories.\\
    & \\
    & Summary: The book "Amber: The Orange Fairy" by Daisy Meadows is a part of the RAINBOW MAGIC series, which is a New York Times bestseller. It falls under the category of Children's Books with a focus on fantasy and humorous stories.\\
    & \\
    & Question: Which category does this book belong to? Please directly give the most likely answer from the following categories: "Literature \& Fiction", "Animals", "Growing Up \& Facts of Life", "Humor", "Cars, Trains \& Things That Go", "Fairy Tales, Folk Tales \& Myths", "Activities, Crafts \& Games", "Science Fiction \& Fantasy", "Classics", "Mysteries \& Detectives", "Action \& Adventure", "Geography \& Cultures", "Education \& Reference", "Arts, Music \& Photography", "Holidays \& Celebrations", "Science, Nature \& How It Works", "Early Learning", "Biographies", "History", "Children's Cookbooks", "Religions", "Sports \& Outdoors", "Comics \& Graphic Novels", "Computers \& Technology". ASSISTANT: \\
\bottomrule
\end{longtable}
In this section, we illustrate the prompt design used in our framework in Table~7, taking the Children dataset as a representative example. For each type of prompt, we describe its role in the reasoning-acting process and provide a concrete instance. For other datasets, the prompt structure remains the same, while dataset-specific descriptions are adjusted accordingly.

\section{Limitations} \label{app.sec.limitation}
Despite its effectiveness, \model\ has several limitations.  First, the framework introduces additional computational overhead due to multi-step LLM inference.  Second, its performance may depend on the availability and quality of node-associated text. Finally, the current design is constrained by the input length of the LLM, which limits the number of reasoning steps and retrieved evidence. 
We leave improving efficiency and scalability for future work.

\section{Broader Impact}
This work introduces a graph-aware reasoning framework that integrates large language models with structured data, which may benefit a wide range of applications involving relational reasoning, such as recommendation systems, knowledge discovery, and scientific data analysis. By enabling zero-shot generalization across graphs, our approach may reduce the need for extensive labeled data and improve the accessibility of graph learning techniques in real-world scenarios.
Potential risks are limited but may arise if the method is applied to sensitive domains, where incorrect predictions or biases inherited from pre-trained language models could propagate through the reasoning process. However, our framework does not introduce new data sources or amplify such biases beyond those already present in existing models.
Overall, we believe this work has a positive impact by advancing the integration of structured and unstructured knowledge for more flexible and generalizable machine learning systems.


\newpage
\section*{NeurIPS Paper Checklist}

\begin{enumerate}

\item {\bf Claims}
    \item[] Question: Do the main claims made in the abstract and introduction accurately reflect the paper's contributions and scope?
    \item[] Answer: \answerYes{} 
    \item[] Justification: The abstract and introduction clearly state the main contributions of the paper, including the proposed reasoning-acting framework, the design of graph-aware retrieval and context refinement, and the focus on zero-shot graph learning. These claims are consistently supported by the methodological descriptions and experimental results presented in the paper, without overstating the scope or generality.
    \item[] Guidelines:
    \begin{itemize}
        \item The answer \answerNA{} means that the abstract and introduction do not include the claims made in the paper.
        \item The abstract and/or introduction should clearly state the claims made, including the contributions made in the paper and important assumptions and limitations. A \answerNo{} or \answerNA{} answer to this question will not be perceived well by the reviewers. 
        \item The claims made should match theoretical and experimental results, and reflect how much the results can be expected to generalize to other settings. 
        \item It is fine to include aspirational goals as motivation as long as it is clear that these goals are not attained by the paper. 
    \end{itemize}

\item {\bf Limitations}
    \item[] Question: Does the paper discuss the limitations of the work performed by the authors?
    \item[] Answer: \answerYes{} 
    \item[] Justification: In Sect.~\ref{app.sec.limitation}, we introduce the limitation of \model.
    \item[] Guidelines:
    \begin{itemize}
        \item The answer \answerNA{} means that the paper has no limitation while the answer \answerNo{} means that the paper has limitations, but those are not discussed in the paper. 
        \item The authors are encouraged to create a separate ``Limitations'' section in their paper.
        \item The paper should point out any strong assumptions and how robust the results are to violations of these assumptions (e.g., independence assumptions, noiseless settings, model well-specification, asymptotic approximations only holding locally). The authors should reflect on how these assumptions might be violated in practice and what the implications would be.
        \item The authors should reflect on the scope of the claims made, e.g., if the approach was only tested on a few datasets or with a few runs. In general, empirical results often depend on implicit assumptions, which should be articulated.
        \item The authors should reflect on the factors that influence the performance of the approach. For example, a facial recognition algorithm may perform poorly when image resolution is low or images are taken in low lighting. Or a speech-to-text system might not be used reliably to provide closed captions for online lectures because it fails to handle technical jargon.
        \item The authors should discuss the computational efficiency of the proposed algorithms and how they scale with dataset size.
        \item If applicable, the authors should discuss possible limitations of their approach to address problems of privacy and fairness.
        \item While the authors might fear that complete honesty about limitations might be used by reviewers as grounds for rejection, a worse outcome might be that reviewers discover limitations that aren't acknowledged in the paper. The authors should use their best judgment and recognize that individual actions in favor of transparency play an important role in developing norms that preserve the integrity of the community. Reviewers will be specifically instructed to not penalize honesty concerning limitations.
    \end{itemize}

\item {\bf Theory assumptions and proofs}
    \item[] Question: For each theoretical result, does the paper provide the full set of assumptions and a complete (and correct) proof?
    \item[] Answer: \answerNA{} 
    \item[] Justification: This paper does not include formal theoretical results such as theorems or proofs. The contributions are primarily methodological and empirical, focusing on the design of a reasoning-acting framework and its experimental evaluation.
\item[] Guidelines:
    \begin{itemize}
        \item The answer \answerNA{} means that the paper does not include theoretical results. 
        \item All the theorems, formulas, and proofs in the paper should be numbered and cross-referenced.
        \item All assumptions should be clearly stated or referenced in the statement of any theorems.
        \item The proofs can either appear in the main paper or the supplemental material, but if they appear in the supplemental material, the authors are encouraged to provide a short proof sketch to provide intuition. 
        \item Inversely, any informal proof provided in the core of the paper should be complemented by formal proofs provided in appendix or supplemental material.
        \item Theorems and Lemmas that the proof relies upon should be properly referenced. 
    \end{itemize}

    \item {\bf Experimental result reproducibility}
    \item[] Question: Does the paper fully disclose all the information needed to reproduce the main experimental results of the paper to the extent that it affects the main claims and/or conclusions of the paper (regardless of whether the code and data are provided or not)?
    \item[] Answer: \answerYes{} 
    \item[] Justification: We provide sufficient details for reproducing the experiments, including model architecture, training and inference procedures, and hyperparameter settings. In addition, we release the source code to facilitate verification and replication of our results.
    \item[] Guidelines:
    \begin{itemize}
        \item The answer \answerNA{} means that the paper does not include experiments.
        \item If the paper includes experiments, a \answerNo{} answer to this question will not be perceived well by the reviewers: Making the paper reproducible is important, regardless of whether the code and data are provided or not.
        \item If the contribution is a dataset and\slash or model, the authors should describe the steps taken to make their results reproducible or verifiable. 
        \item Depending on the contribution, reproducibility can be accomplished in various ways. For example, if the contribution is a novel architecture, describing the architecture fully might suffice, or if the contribution is a specific model and empirical evaluation, it may be necessary to either make it possible for others to replicate the model with the same dataset, or provide access to the model. In general. releasing code and data is often one good way to accomplish this, but reproducibility can also be provided via detailed instructions for how to replicate the results, access to a hosted model (e.g., in the case of a large language model), releasing of a model checkpoint, or other means that are appropriate to the research performed.
        \item While NeurIPS does not require releasing code, the conference does require all submissions to provide some reasonable avenue for reproducibility, which may depend on the nature of the contribution. For example
        \begin{enumerate}
            \item If the contribution is primarily a new algorithm, the paper should make it clear how to reproduce that algorithm.
            \item If the contribution is primarily a new model architecture, the paper should describe the architecture clearly and fully.
            \item If the contribution is a new model (e.g., a large language model), then there should either be a way to access this model for reproducing the results or a way to reproduce the model (e.g., with an open-source dataset or instructions for how to construct the dataset).
            \item We recognize that reproducibility may be tricky in some cases, in which case authors are welcome to describe the particular way they provide for reproducibility. In the case of closed-source models, it may be that access to the model is limited in some way (e.g., to registered users), but it should be possible for other researchers to have some path to reproducing or verifying the results.
        \end{enumerate}
    \end{itemize}

\item {\bf Open access to data and code}
    \item[] Question: Does the paper provide open access to the data and code, with sufficient instructions to faithfully reproduce the main experimental results, as described in supplemental material?
    \item[] Answer: \answerYes{} 
    \item[] Justification: We provide an anonymized code repository with detailed instructions for reproducing the experiments, including environment setup, data preparation, and execution commands. The link to the code is included in the abstract and supplemental material.
    \item[] Guidelines:
    \begin{itemize}
        \item The answer \answerNA{} means that paper does not include experiments requiring code.
        \item Please see the NeurIPS code and data submission guidelines (\url{https://neurips.cc/public/guides/CodeSubmissionPolicy}) for more details.
        \item While we encourage the release of code and data, we understand that this might not be possible, so \answerNo{} is an acceptable answer. Papers cannot be rejected simply for not including code, unless this is central to the contribution (e.g., for a new open-source benchmark).
        \item The instructions should contain the exact command and environment needed to run to reproduce the results. See the NeurIPS code and data submission guidelines (\url{https://neurips.cc/public/guides/CodeSubmissionPolicy}) for more details.
        \item The authors should provide instructions on data access and preparation, including how to access the raw data, preprocessed data, intermediate data, and generated data, etc.
        \item The authors should provide scripts to reproduce all experimental results for the new proposed method and baselines. If only a subset of experiments are reproducible, they should state which ones are omitted from the script and why.
        \item At submission time, to preserve anonymity, the authors should release anonymized versions (if applicable).
        \item Providing as much information as possible in supplemental material (appended to the paper) is recommended, but including URLs to data and code is permitted.
    \end{itemize}

\item {\bf Experimental setting/details}
    \item[] Question: Does the paper specify all the training and test details (e.g., data splits, hyperparameters, how they were chosen, type of optimizer) necessary to understand the results?
    \item[] Answer: \answerYes{} 
    \item[] Justification: We provide detailed experimental settings in the main paper and appendix, including dataset splits, model configurations, hyperparameters, and inference procedures. Additional implementation details are also included in the released code to ensure clarity and reproducibility.
    \item[] Guidelines:
    \begin{itemize}
        \item The answer \answerNA{} means that the paper does not include experiments.
        \item The experimental setting should be presented in the core of the paper to a level of detail that is necessary to appreciate the results and make sense of them.
        \item The full details can be provided either with the code, in appendix, or as supplemental material.
    \end{itemize}

\item {\bf Experiment statistical significance}
    \item[] Question: Does the paper report error bars suitably and correctly defined or other appropriate information about the statistical significance of the experiments?
    \item[] Answer: \answerYes{} 
    \item[] Justification: We report the mean and standard deviation of performance over five runs with different random seeds. The error bars reflect the variability across different random initializations and data splits, and are presented in both tables and figures. We explicitly indicate that the reported uncertainty corresponds to standard deviation.
    \item[] Guidelines:
    \begin{itemize}
        \item The answer \answerNA{} means that the paper does not include experiments.
        \item The authors should answer \answerYes{} if the results are accompanied by error bars, confidence intervals, or statistical significance tests, at least for the experiments that support the main claims of the paper.
        \item The factors of variability that the error bars are capturing should be clearly stated (for example, train/test split, initialization, random drawing of some parameter, or overall run with given experimental conditions).
        \item The method for calculating the error bars should be explained (closed form formula, call to a library function, bootstrap, etc.)
        \item The assumptions made should be given (e.g., Normally distributed errors).
        \item It should be clear whether the error bar is the standard deviation or the standard error of the mean.
        \item It is OK to report 1-sigma error bars, but one should state it. The authors should preferably report a 2-sigma error bar than state that they have a 96\% CI, if the hypothesis of Normality of errors is not verified.
        \item For asymmetric distributions, the authors should be careful not to show in tables or figures symmetric error bars that would yield results that are out of range (e.g., negative error rates).
        \item If error bars are reported in tables or plots, the authors should explain in the text how they were calculated and reference the corresponding figures or tables in the text.
    \end{itemize}

\item {\bf Experiments compute resources}
    \item[] Question: For each experiment, does the paper provide sufficient information on the computer resources (type of compute workers, memory, time of execution) needed to reproduce the experiments?
    \item[] Answer: \answerYes{} 
    \item[] Justification: We provide the type of compute resources (GPU models) and memory specifications used in our experiments. While we do not explicitly report the execution time, the provided information is sufficient to reproduce the experimental setup and results.
    \item[] Guidelines:
    \begin{itemize}
        \item The answer \answerNA{} means that the paper does not include experiments.
        \item The paper should indicate the type of compute workers CPU or GPU, internal cluster, or cloud provider, including relevant memory and storage.
        \item The paper should provide the amount of compute required for each of the individual experimental runs as well as estimate the total compute. 
        \item The paper should disclose whether the full research project required more compute than the experiments reported in the paper (e.g., preliminary or failed experiments that didn't make it into the paper). 
    \end{itemize}
    
\item {\bf Code of ethics}
    \item[] Question: Does the research conducted in the paper conform, in every respect, with the NeurIPS Code of Ethics \url{https://neurips.cc/public/EthicsGuidelines}?
    \item[] Answer: \answerYes{} 
    \item[] Justification: The research presented in this paper fully complies with the NeurIPS Code of Ethics. We use only publicly available datasets, ensure proper citation of prior work, and maintain anonymity in the submission. No human subjects or sensitive data are involved.
    \item[] Guidelines:
    \begin{itemize}
        \item The answer \answerNA{} means that the authors have not reviewed the NeurIPS Code of Ethics.
        \item If the authors answer \answerNo, they should explain the special circumstances that require a deviation from the Code of Ethics.
        \item The authors should make sure to preserve anonymity (e.g., if there is a special consideration due to laws or regulations in their jurisdiction).
    \end{itemize}

\item {\bf Broader impacts}
    \item[] Question: Does the paper discuss both potential positive societal impacts and negative societal impacts of the work performed?
    \item[] Answer: \answerYes{} 
    \item[] Justification: In appendix, we discuss the broader impacts of \model.
    \item[] Guidelines:
    \begin{itemize}
        \item The answer \answerNA{} means that there is no societal impact of the work performed.
        \item If the authors answer \answerNA{} or \answerNo, they should explain why their work has no societal impact or why the paper does not address societal impact.
        \item Examples of negative societal impacts include potential malicious or unintended uses (e.g., disinformation, generating fake profiles, surveillance), fairness considerations (e.g., deployment of technologies that could make decisions that unfairly impact specific groups), privacy considerations, and security considerations.
        \item The conference expects that many papers will be foundational research and not tied to particular applications, let alone deployments. However, if there is a direct path to any negative applications, the authors should point it out. For example, it is legitimate to point out that an improvement in the quality of generative models could be used to generate Deepfakes for disinformation. On the other hand, it is not needed to point out that a generic algorithm for optimizing neural networks could enable people to train models that generate Deepfakes faster.
        \item The authors should consider possible harms that could arise when the technology is being used as intended and functioning correctly, harms that could arise when the technology is being used as intended but gives incorrect results, and harms following from (intentional or unintentional) misuse of the technology.
        \item If there are negative societal impacts, the authors could also discuss possible mitigation strategies (e.g., gated release of models, providing defenses in addition to attacks, mechanisms for monitoring misuse, mechanisms to monitor how a system learns from feedback over time, improving the efficiency and accessibility of ML).
    \end{itemize}
    
\item {\bf Safeguards}
    \item[] Question: Does the paper describe safeguards that have been put in place for responsible release of data or models that have a high risk for misuse (e.g., pre-trained language models, image generators, or scraped datasets)?
    \item[] Answer: \answerNA{} 
    \item[] Justification: This work does not introduce or release new datasets or models with high risk of misuse. We use publicly available datasets and existing pre-trained models, and therefore no additional safeguards are required.
    \item[] Guidelines:
    \begin{itemize}
        \item The answer \answerNA{} means that the paper poses no such risks.
        \item Released models that have a high risk for misuse or dual-use should be released with necessary safeguards to allow for controlled use of the model, for example by requiring that users adhere to usage guidelines or restrictions to access the model or implementing safety filters. 
        \item Datasets that have been scraped from the Internet could pose safety risks. The authors should describe how they avoided releasing unsafe images.
        \item We recognize that providing effective safeguards is challenging, and many papers do not require this, but we encourage authors to take this into account and make a best faith effort.
    \end{itemize}

\item {\bf Licenses for existing assets}
    \item[] Question: Are the creators or original owners of assets (e.g., code, data, models), used in the paper, properly credited and are the license and terms of use explicitly mentioned and properly respected?
    \item[] Answer: \answerYes{} 
    \item[] Justification: We properly cite all datasets and models used in this work and provide corresponding references and URLs. These datasets are publicly available and widely used in prior literature, and we follow their standard usage and licensing terms.
    \item[] Guidelines:
    \begin{itemize}
        \item The answer \answerNA{} means that the paper does not use existing assets.
        \item The authors should cite the original paper that produced the code package or dataset.
        \item The authors should state which version of the asset is used and, if possible, include a URL.
        \item The name of the license (e.g., CC-BY 4.0) should be included for each asset.
        \item For scraped data from a particular source (e.g., website), the copyright and terms of service of that source should be provided.
        \item If assets are released, the license, copyright information, and terms of use in the package should be provided. For popular datasets, \url{paperswithcode.com/datasets} has curated licenses for some datasets. Their licensing guide can help determine the license of a dataset.
        \item For existing datasets that are re-packaged, both the original license and the license of the derived asset (if it has changed) should be provided.
        \item If this information is not available online, the authors are encouraged to reach out to the asset's creators.
    \end{itemize}

\item {\bf New assets}
    \item[] Question: Are new assets introduced in the paper well documented and is the documentation provided alongside the assets?
    \item[] Answer: \answerYes{} 
    \item[] Justification:  We release the code for our method and provide detailed documentation, including a README file with instructions for setup, data preparation, and reproducing the experiments.
    \item[] Guidelines:
    \begin{itemize}
        \item The answer \answerNA{} means that the paper does not release new assets.
        \item Researchers should communicate the details of the dataset\slash code\slash model as part of their submissions via structured templates. This includes details about training, license, limitations, etc. 
        \item The paper should discuss whether and how consent was obtained from people whose asset is used.
        \item At submission time, remember to anonymize your assets (if applicable). You can either create an anonymized URL or include an anonymized zip file.
    \end{itemize}

\item {\bf Crowdsourcing and research with human subjects}
    \item[] Question: For crowdsourcing experiments and research with human subjects, does the paper include the full text of instructions given to participants and screenshots, if applicable, as well as details about compensation (if any)? 
    \item[] Answer: \answerNA{} 
    \item[] Justification: This work does not involve crowdsourcing or research with human subjects.
    \item[] Guidelines:
    \begin{itemize}
        \item The answer \answerNA{} means that the paper does not involve crowdsourcing nor research with human subjects.
        \item Including this information in the supplemental material is fine, but if the main contribution of the paper involves human subjects, then as much detail as possible should be included in the main paper. 
        \item According to the NeurIPS Code of Ethics, workers involved in data collection, curation, or other labor should be paid at least the minimum wage in the country of the data collector. 
    \end{itemize}

\item {\bf Institutional review board (IRB) approvals or equivalent for research with human subjects}
    \item[] Question: Does the paper describe potential risks incurred by study participants, whether such risks were disclosed to the subjects, and whether Institutional Review Board (IRB) approvals (or an equivalent approval/review based on the requirements of your country or institution) were obtained?
    \item[] Answer: \answerNA{} 
    \item[] Justification: The paper does not involve crowdsourcing nor research with human subjects
    \item[] Guidelines:
    \begin{itemize}
        \item The answer \answerNA{} means that the paper does not involve crowdsourcing nor research with human subjects.
        \item Depending on the country in which research is conducted, IRB approval (or equivalent) may be required for any human subjects research. If you obtained IRB approval, you should clearly state this in the paper. 
        \item We recognize that the procedures for this may vary significantly between institutions and locations, and we expect authors to adhere to the NeurIPS Code of Ethics and the guidelines for their institution. 
        \item For initial submissions, do not include any information that would break anonymity (if applicable), such as the institution conducting the review.
    \end{itemize}

\item {\bf Declaration of LLM usage}
    \item[] Question: Does the paper describe the usage of LLMs if it is an important, original, or non-standard component of the core methods in this research? Note that if the LLM is used only for writing, editing, or formatting purposes and does \emph{not} impact the core methodology, scientific rigor, or originality of the research, declaration is not required.
    \item[] Answer: \answerNA{} 
    \item[] Justification: The core method development in this research does not involve LLMs as any important, original, or non-standard components.
    \item[] Guidelines:
    \begin{itemize}
        \item The answer \answerNA{} means that the core method development in this research does not involve LLMs as any important, original, or non-standard components.
        \item Please refer to our LLM policy in the NeurIPS handbook for what should or should not be described.
    \end{itemize}

\end{enumerate}

\end{document}